\documentclass[10pt,twoside]{siamltex}
\usepackage[dvips]{graphicx} 
\usepackage{epsfig}  
\usepackage{algorithmic}
\usepackage{algorithm}
\usepackage{textcomp}
\usepackage{verbatim}
\usepackage{amsmath}
\usepackage{amsfonts}
\usepackage{amssymb}
\usepackage{epsfig}  
\usepackage{subfigure}
\DeclareMathOperator*{\nut}{arg}

\begin{document}



\bibliographystyle{plain}
\title{
Clustering and Latent Semantic Indexing Aspects of the Nonnegative Matrix Factorization}

\author{
Andri Mirzal\thanks{Faculty of Computer Science and Information Systems, University of Technology Malaysia, (andrimirzal@utm.my.}}

\pagestyle{myheadings}
\markboth{A.~Mirzal}{Clustering and LSI aspects of the NMF}
\maketitle

\begin{abstract}
This paper provides a theoretical support for clustering aspect of the nonnegative matrix factorization (NMF). By utilizing the Karush-Kuhn-Tucker optimality conditions, we show that NMF objective is equivalent to graph clustering objective, so clustering aspect of the NMF has a solid justification. Different from previous approaches which usually discard the nonnegativity constraints, our approach guarantees the stationary point being used in deriving the equivalence is located on the feasible region in the nonnegative orthant. Additionally, since clustering capability of a matrix decomposition technique can sometimes imply its latent semantic indexing (LSI) aspect, we will also evaluate LSI aspect of the NMF by showing its capability in solving the synonymy and polysemy problems in synthetic datasets. And more extensive evaluation will be conducted by comparing LSI performances of the NMF and the singular value decomposition (SVD)---the standard LSI method---using some standard datasets.
\end{abstract}


\section{Introduction} \label{intro}

Nonnegative datasets are everywhere; from term by document matrix induced from a document corpus \cite{Xu,Shahnaz}, gene expression datasets \cite{HKim2}, pixels in digital images \cite{Lee}, disease patterns \cite{Hoyer}, to spectral signatures from astronomical spectrometers \cite{Pauca} among others. Even though diverse, they have one thing in common: all can be represented by using nonnegative matrices induced from the datasets. This allows many well-established mathematical techniques to be applied in order to anayze the datasets.

There are many common tasks associated with these datasets, for example: grouping the similar data points (\emph{clustering}), finding patterns in the datasets, identifying important or interesting features, and finding sets of relevant data points to queries (\emph{information retrieval}). In this paper, we will focus on two tasks: clustering and latent semantic indexing---a technique that can be used for improving recall and precision of an information retrieval (IR) system.

\subsection{Clustering}

Clustering is the task of assigning data points into clusters such that similar points are in the same clusters and dissimilar points are in the dif\mbox{}ferent clusters. There are many types of clustering, for example supervised/unsu-pervised, hierarchical/partitional, hard/soft, and one-way/many-way (two-way clustering is known as co-clustering or bi-clustering) among others. In this paper, clustering term refers to unsupervised, partitional, hard, and one-way clustering. Further, the number of cluster is given beforehand. 

The NMF as a clustering method can be traced back to the work by Lee \& Seung \cite{Lee}. But, the first work that explicitly demonstrates it is the work by Xu et al.~\cite{Xu} in which they show that the NMF outperforms the spectral methods in term of \emph{purity} and \emph{mutual information} measures for Reuters and TDT2 datasets. 

Clustering aspect of the NMF, even though numerically well studied, is not theoretically well explained. Usually this aspect is explained by showing the equivalence between NMF objective to either k-means clustering objective \cite{Ding2, JKim} or spectral clustering objective \cite{Ding2}. The problem with the first approach is there is no obvious way to incorporate the nonnegativity constraints into k-means clustering objective. And the problem with the second approach is it discards the nonnegativity constraints, thus is equivalent to finding stationary points on unbounded region. Accordingly, the NMF which is a bound-constrained optimization turns into an unbounded optimization, so there is no guarantee the stationary point being utilized in proving the equivalence is located on the feasible region indicated by the constraints.

In the first part of this paper, we will provide a theoretical support for clustering aspect of the NMF by analyzing the objective at the stationary point using the Karush-Kuhn-Tucker (KKT) conditions without setting the KKT multipliers to zeros. Thus, the stationary point under investigation is guaranteed to be located on the feasible region.

\subsection{Latent semantic indexing}

Latent semantic indexing (LSI) is a method introduced by Deerwester et al.~\cite{Deerwester} to improve recall and precision of an IR system using truncated singular value decomposition (SVD) of the term-by-document matrix to reveal hidden relationship between documents by indexing terms that are present in the similar documents and weakening the influences of terms that are mutually present in the dissimilar documents. The first capability can solve the \emph{synonymy}---dif\mbox{}ferent words with similar meaning---problem, and the second capability can solve the \emph{polysemy}---words with multiple unrelated meanings---problem. Thus, LSI not only is able to retrieve relevant documents that do not contain terms in the query, but also can filter out irrelevant documents that contain terms in the query.

LSI aspect of the NMF is not well studied. There are some works that discuss the relationship between the NMF and probabilistic LSI, e.g., \cite{Gaussier, Dingz}. But the emphasize is in clustering capability of probabilistic LSI, not LSI aspect of the NMF. Motivated by the SVD which is the standard method in clustering and LSI, in the second part of this paper, LSI aspect of the NMF will be studied, and the results will be compared to the results of the SVD.

\section{The nonnegative matrix factorization} \label{nmf}

The NMF was popularized by the work of Lee \& Seung \cite{Lee} in which they showed that this technique can be used to learn parts of faces and semantic features of text. Previously, it has been studied under the term positive matrix factorization \cite{Paatero, Anttila}. Mathematically, the NMF is a technique that decomposes a nonnegative data matrix into a pair of other nonnegative matrices:
\begin{equation}
\mathbf{A} \approx \mathbf{B}\mathbf{C},
\label{ch3:eq1}
\end{equation}
where $\mathbf{A}\in\mathbb{R}_{+}^{M\times N}=\left[\mathbf{a}_1,\ldots,\mathbf{a}_N\right]$ denotes the data matrix, $\mathbf{B}\in\mathbb{R}_{+}^{M\times K}=\left[\mathbf{b}_1,\ldots,\mathbf{b}_K\right]$ denotes the basis matrix, $\mathbf{C}\in\mathbb{R}_{+}^{K\times N}=\left[\mathbf{c}_1,\ldots,\mathbf{c}_N\right]$ denotes the coef\mbox{}ficient matrix, and $K$ denotes the number of factors which usually is chosen so that $K\ll\min(M,N)$. Note that the definitions of $\mathbf{A}$, $\mathbf{B}$, and $\mathbf{C}$ are chosen to simplify the interpretation of the NMF. 

To compute $\mathbf{B}$ and $\mathbf{C}$, usually eq.~\ref{ch3:eq1} is rewritten into a minimization problem in Frobenius norm.
\begin{equation}
\min_{\mathbf{B},\mathbf{C}}J(\mathbf{B},\mathbf{C})=\frac{1}{2}\|\mathbf{A}-\mathbf{B}\mathbf{C}\|_{F}^{2}\;\,\mathrm{s.t.}\;\, \mathbf{B}\ge\mathbf{0},\mathbf{C}\ge\mathbf{0}.
\label{ch3:eq2}
\end{equation}
In addition to the usual Frobenius norm, family of Bregman divergences---which Frobenius norm and Kullback-Leibler divergence are part of it---can also be used as the distance measures. Detailed discussion on Bregman divergences can be found in, e.g., ref.~\cite{Dhillon1}. In this work, we will consider only Frobenius norm.

\section{Limit points of the sequences generated by NMF algorithms} \label{ch3:limitpoint}

All NMF algorithms are formulated in the alternating fashion, fixing one matrix while solving the other (the popular Lee \& Seung algorithms \cite{Lee2} and their derivatives, e.g., \cite{Xu, Hoyer, SZLi, CJLin2} also use the alternating strategy, but cannot be represented by generic algorithm below). This strategy is employed because the NMF is nonconvex with respect to $\mathbf{B}$ and $\mathbf{C}$, but is convex with respect to $\mathbf{B}$ or $\mathbf{C}$ \cite{CJLin}. Thus, the alternating strategy transforms NMF problem into a pair of convex subproblems. Transforming a nonconvex problem into the corresponding convex subproblems is a common practice in optimization researches because: (1) convex optimization is more tractable, (2) usually convex methods are more ef\mbox{}ficient, (3) any local optimum is necessarily a global optimum, and (4) the algorithms are easy to initialize \cite{Hindi}.
\begin{algorithm}
\caption{Generic algorithm for the NMF based on ANLS.}
\label{ch3:alg1}
\begin{algorithmic}
\STATE Initialization: $\mathbf{C}^0\ge\mathbf{0}$.
\FOR {$l=0,\ldots$}
\STATE \begin{align}
\mathbf{B}^{(l+1)} &\longleftarrow \nut_{\mathbf{B}\ge\mathbf{0}}\min\frac{1}{2}\|\mathbf{A}-\mathbf{B}\mathbf{C}^{(l)}\|_F^2 \label{ch3:eq3}\\
\mathbf{C}^{(l+1)} &\longleftarrow \nut_{\mathbf{C}\ge\mathbf{0}}\min\frac{1}{2}\|\mathbf{A}-\mathbf{B}^{(l+1)}\mathbf{C}\|_F^2 \label{ch3:eq4},
\end{align}
\ENDFOR
\end{algorithmic}
\end{algorithm}

Algorithm \ref{ch3:alg1} solves NMF problem in the alternating fashion which will generate a solution sequence $\{\mathbf{B}^{(l)},\mathbf{C}^{(l)}\}_{l=0}^{L}$. This algorithm is known as alternating nonnegativity-constrained least square (ANLS) algorithm, and usually is solved by decomposing each subproblem into the corresponding nonnegativity-constrained least square (NNLS) problems, where there are many algorithms that guarantee the global-optimality of the NNLS problems. The following equations are the NNLS versions of the ANLS in algorithm \ref{ch3:alg1}:
\begin{align}
\mathbf{\hat{b}}_m^{T(l+1)} &\longleftarrow \nut_{\mathbf{\hat{b}}_m^T\ge\mathbf{0}}\min\frac{1}{2}\|\mathbf{\hat{a}}_m^T-\mathbf{C}^{T(l)}\mathbf{\hat{b}}_m^T\|_F^2,\;\; \forall m \label{ch3:eq5}\\
\mathbf{c}_n^{(l+1)} &\longleftarrow \nut_{\mathbf{c}_n\ge\mathbf{0}}\min\frac{1}{2}\|\mathbf{a}_n-\mathbf{B}^{(l+1)}\mathbf{c}_n\|_F^2,\;\; \forall n \label{ch3:eq6},
\end{align}
where $\mathbf{\hat{x}}_i$ is the $i$-th row of $\mathbf{X}$.

According to Grippo \& Sciandrone \cite{Grippo}, any limit point of $\{\mathbf{B}^{(l)}$, $\mathbf{C}^{(l)}\}_{l=0}^L$---generated by any ANLS algorithm that optimally solves the convex subproblem eq.~\ref{ch3:eq3} and eq.~\ref{ch3:eq4}---is a stationary point. And such ANLS based NMF algorithms exist, e.g., \cite{CJLin2, CJLin, DKim, DKim2, HKim, JKim2}, therefore there is guarantee that the stationary points are reachable. And as NNLS is the building block for ANLS, any NNLS algorithm that guarantees to find optimal solutions of eq.~\ref{ch3:eq5} and eq.~\ref{ch3:eq6}, e.g., \cite{Benthem, Bro, Lawson} can also be employed to search for the stationary points. And as will be shown in section \ref{ch3:clusteringnmf}, NMF objective (eq.~\ref{ch3:eq2}) implicitly puts upper bounds on the feasible region (the lower bounds are explicit: the nonnegativity constraints). Thus the NMF is bound-constrained optimization problem, consequently $\{\mathbf{B}^{(l)}$, $\mathbf{C}^{(l)}\}_{l=0}^L$ has at least one limit point \cite{CJLin}. This completes the conditions for any NMF algorithm that optimally solves subproblem eq.~\ref{ch3:eq3} and eq.~\ref{ch3:eq4} to have convergence guarantee.

\section{Clustering aspect of the NMF} \label{ch3:clusteringnmf}

This section is the first part of this paper in which a theoretical framework for supporting clustering aspect of the NMF will be provided. The strict KKT optimality conditions will be utilized to derive the equivalence between NMF objective to graph clustering objective. Unlike previous approaches where the KKT multipliers are set to zeros \cite{Ding2, TLi, Ding0, Ding1}, we will make no assumption about the KKT multipliers, thus the stationary point under investigation is guaranteed to be located on the feasible region in the nonnegative orthant.

We will also show that the feasible region is bounded, with the lower bounds are explicitly bounded by the nonnegativity constraints, and the upper bounds are implicitly bounded by the objective. As stated in section \ref{ch3:limitpoint}, the boundedness of the feasible region is the necessary condition for guaranteeing the existence of limit point of $\{\mathbf{B}^{(l)}, \mathbf{C}^{(l)}\}_{l=0}^L$. And for interpretability reason, the data matrix $\mathbf{A}$ will be considered as a feature-by-item data matrix unless stated dif\mbox{}ferently.

The following proposition gives the theoretical support for clustering aspect of the NMF.
\begin{proposition} \label{ch3:prop1}
Minimizing the following objective
\begin{align}
&\min_{\mathbf{B},\mathbf{C}}J(\mathbf{B},\mathbf{C})=\frac{1}{2}\|\mathbf{A}-\mathbf{BC}\|_{F}^{2} \label{ch3:eq9}\\
&\mathrm{s.t.}\;\,\mathbf{B}\ge\mathbf{0},\mathbf{C}\ge\mathbf{0}, \nonumber
\end{align}
leads to the feature clustering indicator matrix $\mathbf{B}$ and the item clustering indicator matrix $\mathbf{C}$.
\end{proposition}
\begin{proof}
\begin{equation*}
\|\mathbf{A}-\mathbf{B}\mathbf{C}\|_{F}^{2} = \mathrm{tr}\;(\mathbf{A}^T\mathbf{A}-2\mathbf{CA}^T\mathbf{B}+\mathbf{B}^T\mathbf{BCC}^T).
\end{equation*}
Since $\mathbf{A}$ is constant, minimizing $J$ is equivalent to simultaneously optimizing:
\begin{align}
&\max_{\mathbf{B},\mathbf{C}}\mathrm{tr}\;(\mathbf{CA}^T\mathbf{B}) \label{ch3:eq10} \\
&\min_{\mathbf{B},\mathbf{C}}\mathrm{tr}\;(\mathbf{B}^T\mathbf{BCC}^T). \label{ch3:eq11}
\end{align}
Note that because $\mathrm{tr}\,(\mathbf{XY})\le\mathrm{tr}\,(\mathbf{X})\;\mathrm{tr}\,(\mathbf{Y})$, minimizing Eq.~\ref{ch3:eq11} is equivalent to:
\begin{align}
&\min_{\mathbf{B}}\mathrm{tr}\;(\mathbf{B}^T\mathbf{B}) \label{ch3:eq12}\;\, \text{and} \\
&\min_{\mathbf{C}}\mathrm{tr}\;(\mathbf{CC}^T). \label{ch3:eq13}
\end{align}

The KKT function of objective in eq.~\ref{ch3:eq9} is:
\begin{equation*}
L(\mathbf{B},\mathbf{C})=\;J(\mathbf{B},\mathbf{C})-\mathrm{tr}\;(\mathbf{\Gamma}_{\mathbf{B}}\mathbf{B}^T)-\mathrm{tr}\;(\mathbf{\Gamma}_{\mathbf{C}}\mathbf{C}), 
\end{equation*}
where $\mathbf{\Gamma}_{\mathbf{B}}\in\mathbb{R}_{+}^{M\times K}$ and $\mathbf{\Gamma}_{\mathbf{C}}\in\mathbb{R}_{+}^{N\times K}$ are the KKT multipliers. By applying the KKT optimality conditions to $L$ we get:
\begin{align}
\nabla_{\mathbf{B}}L=\;&\mathbf{BCC}^T-\mathbf{AC}^T-\mathbf{\Gamma}_{\mathbf{B}}=\mathbf{0} \label{ch3:eq15} \\
\nabla_{\mathbf{C}}L=\;&\mathbf{B}^T\mathbf{BC}-\mathbf{B}^T\mathbf{A}-\mathbf{\Gamma}_{\mathbf{C}}^T=\mathbf{0},\label{ch3:eq16}
\end{align}
with complementary slackness:
\begin{equation*}
\mathbf{\Gamma}_{\mathbf{B}}\odot\mathbf{B}=\mathbf{0},\;\,\text{and}\;\,\mathbf{\Gamma}_{\mathbf{C}}^T\odot\mathbf{C}=\mathbf{0}, 
\end{equation*}
where $\odot$ denotes component-wise multiplications. Eq.~\ref{ch3:eq15} and eq.~\ref{ch3:eq16} lead to:
\begin{align}
\mathbf{B} &= (\mathbf{AC}^T+\mathbf{\Gamma}_{\mathbf{B}})(\mathbf{CC}^T)^{^{-1}} \label{ch3:eq17}\\
\mathbf{C} &= (\mathbf{B}^T\mathbf{B})^{^{-1}}(\mathbf{B}^T\mathbf{A}+\mathbf{\Gamma}_{\mathbf{C}}^T). \label{ch3:eq18}
\end{align}
Substituting eq.~\ref{ch3:eq18} to eq.~\ref{ch3:eq10} leads to:
\begin{equation*}
\max_{\mathbf{B}}\mathrm{tr}\;\big((\mathbf{B}^T\mathbf{B})^{^{-1}}(\mathbf{B}^T\mathbf{AA}^T\mathbf{B}+\mathbf{\Gamma}_{\mathbf{C}}^T\mathbf{A}^T\mathbf{B})\big), 
\end{equation*}
which is equivalent to simultaneously optimizing:
\begin{align}
&\max_{\mathbf{B}}\mathrm{tr}\;(\mathbf{B}^T\mathbf{AA}^T\mathbf{B}) \label{ch3:eq20} \\
&\max_{\mathbf{B}}\mathrm{tr}\;(\mathbf{\Gamma}_{\mathbf{C}}^T\mathbf{A}^T\mathbf{B}) \label{ch3:eq21} \\
&\min_{\mathbf{B}}\mathrm{tr}\;(\mathbf{B}^T\mathbf{B}). \label{ch3:eq22}
\end{align}
Similarly, substituting eq.~\ref{ch3:eq17} to eq.~\ref{ch3:eq10} leads to:
\begin{equation*}
\max_{\mathbf{C}}\mathrm{tr}\;\big((\mathbf{CA}^T\mathbf{AC}^T+\mathbf{CA}^T\mathbf{\Gamma}_{\mathbf{B}})(\mathbf{CC}^T)^{^{-1}}\big), 
\end{equation*}
which is equivalent to simultaneously optimizing:
\begin{align}
&\max_{\mathbf{C}}\mathrm{tr}\;(\mathbf{CA}^T\mathbf{AC}^T) \label{ch3:eq24} \\
&\max_{\mathbf{C}}\mathrm{tr}\;(\mathbf{CA}^T\mathbf{\Gamma}_{\mathbf{B}}) \label{ch3:eq25} \\
&\min_{\mathbf{C}}\mathrm{tr}\;(\mathbf{CC}^T). \label{ch3:eq26}
\end{align}
As shown, eq.~\ref{ch3:eq22} and eq.~\ref{ch3:eq26} recover eq.~\ref{ch3:eq12} and eq.~\ref{ch3:eq13} respectively, so there is no need to substituting eq.~\ref{ch3:eq17} and eq.~\ref{ch3:eq18} into eq.~\ref{ch3:eq11}.

Now we concentrate on the basis matrix $\mathbf{B}$ first. Eq.~\ref{ch3:eq20} -- \ref{ch3:eq22} give alternative objectives to the original NMF objective that contain only $\mathbf{B}$. Note that if we consider $\mathbf{A}$ to be an af\mbox{}finity matrix induced from bipartite graph $\mathcal{G}(\mathbf{A})$ (which is a reasonable thought since any feature-by-item matrix can be modeled by a bipartite graph), then $\mathcal{G}(\mathbf{AA}^T)$ is the feature graph where edge weights describe the similarity between corresponding vertex pairs. So, eq.~\ref{ch3:eq20} looks like \emph{ratio association} applied to $\mathcal{G}(\mathbf{AA}^T)$. But without orthogonality constraint $\mathbf{B}^T\mathbf{B}=\mathbf{I}$ (which is the part of \emph{ratio association} objective), one can optimize eq.~\ref{ch3:eq20} by setting $\mathbf{B}$ to an infinity matrix. However, this violates eq.~\ref{ch3:eq22} which favours small $\mathbf{B}$. Similarly, one can optimize eq.~\ref{ch3:eq22} by setting $\mathbf{B}$ to a zero matrix. But again, this violates eq.~\ref{ch3:eq20}. Thus, eq.~\ref{ch3:eq20} and eq.~\ref{ch3:eq22} create implicit lower and upper bound constraints on $\mathbf{B}$: $\mathbf{0}\le\mathbf{B}\le\mathbf{\Upsilon}_{\mathbf{B}}$. 

For convenience, eq.~\ref{ch3:eq22} can be restated as: 
\begin{equation}
\min_{\mathbf{B}}\mathrm{tr}\;(\mathbf{B}^T\mathbf{B}) \equiv \min_{\mathbf{B}}\mathrm{tr}\;(\mathbf{B}^T\mathbf{B}\mathbf{B}^T\mathbf{B}). \label{ch3:eq27}
\end{equation}
By using the fact $\mathrm{tr}\;(\mathbf{X}^T\mathbf{X})=\|\mathbf{X}\|_F^2$, eq.~\ref{ch3:eq27} can be rewritten into:
\begin{equation*}
\min_{\mathbf{B}}\;\Big(\left\|\mathbf{B}^T\mathbf{B}\right\|_F^2=\sum_{i}\left(\mathbf{b}_{i}^T\mathbf{b}_{i}\right)^2 + \sum_{i\ne j}\left(\mathbf{b}_{i}^T\mathbf{b}_{j}\right)^2\Big), 
\end{equation*}
Therefore, eq.~\ref{ch3:eq20} -- \ref{ch3:eq22} can be restated as:
\begin{align}
&\max_{\mathbf{B}}\mathrm{tr}\;(\mathbf{B}^T\mathbf{AA}^T\mathbf{B}) \label{ch3:eq29} \\
&\max_{\mathbf{B}}\mathrm{tr}\;(\mathbf{\Gamma}_{\mathbf{C}}^T\mathbf{A}^T\mathbf{B}) \label{ch3:eq30} \\
&\min_{\mathbf{b}}\;\Big( \underbrace{\sum_{i}\left(\mathbf{b}_{i}^T\mathbf{b}_{i}\right)^2}_{j_{b1}} + \underbrace{\sum_{i\ne j}\left(\mathbf{b}_{i}^T\mathbf{b}_{j}\right)^2}_{j_{b2}} \Big) \label{ch3:eq31} \\
&\text{s.t.}\;\,\mathbf{0}\le\mathbf{B}\le\mathbf{\Upsilon}_{\mathbf{B}}. \nonumber
\end{align}
Even though $\mathbf{B}$ is now bounded, since there is no column-orthogonality constraint, maximizing eq.~\ref{ch3:eq29} can be easily done by setting each entry of $\mathbf{B}$ to the corresponding largest possible value (in graph term this means to only create one partition on $\mathcal{G}(\mathbf{A}\mathbf{A}^T)$). But this scenario results in a large value of eq.~\ref{ch3:eq31}, which violates the objective. Similarly, minimizing eq.~\ref{ch3:eq31} to the smallest possible value violates eq.~\ref{ch3:eq29}. Since minimizing $j_{b1}$ implies minimizing $j_{b2}$, but not vice versa, simultaneously optimizing eq.~\ref{ch3:eq29} and eq.~\ref{ch3:eq31} can be done by setting $j_{b2}$ as small as possible and balancing $j_{b1}$ with eq.~\ref{ch3:eq29}. This scenario is the relaxed \emph{ratio association} applied to $\mathcal{G}(\mathbf{A}\mathbf{A}^T)$, and as long as vertices in $\mathcal{G}(\mathbf{A}\mathbf{A}^T)$ are clustered, it leads to the grouping of related features.

The remaining problem is eq.~\ref{ch3:eq30}. Since we know nothing about $\mathbf{\Gamma}_{\mathbf{C}}$, the best bet will be making each entry of $\mathbf{A}^T\mathbf{B}$ as large as possible. This can be done by setting $\mathbf{B}$ to the largest possible values, but this scenario violates eq.~\ref{ch3:eq31}. So, the most reasonable scenario will be making the entries near diagonal region of $\mathbf{A}^T\mathbf{B}$ as large as possible. This can be achieved by using $\mathbf{B}$ from previous discussion. As $\mathbf{B}$ is the feature clustering indicator matrix, multiplying $\mathbf{A}^T$ with $\mathbf{B}$ will result in a matrix that has larger entries near diagonal region, therefore it can be expected that eq.~\ref{ch3:eq30} will have good optimality. Thus simultaneously optimizing eq.~\ref{ch3:eq29} -- \ref{ch3:eq31} leads to the feature clustering indicator matrix $\mathbf{B}$.

By applying the similar approach to the coef\mbox{}ficient matrix $\mathbf{C}$, optimizing eq.~\ref{ch3:eq24} -- \ref{ch3:eq26} is equivalent to optimizing:
\begin{align}
&\max_{\mathbf{C}}\mathrm{tr}\;(\mathbf{CA}^T\mathbf{AC}^T) \label{ch3:eq32} \\
&\max_{\mathbf{C}}\mathrm{tr}\;(\mathbf{CA}^T\mathbf{\Gamma}_{\mathbf{B}}) \label{ch3:eq33} \\
&\min_{\mathbf{\hat{c}}}\;\Big(\sum_{i}\left(\mathbf{\hat{c}}_{i}\mathbf{\hat{c}}_{i}^T\right)^2 + \sum_{i\ne j}\left(\mathbf{\hat{c}}_{i}\mathbf{\hat{c}}_{j}^T\right)^2 \Big) \label{ch3:eq34} \\
&\text{s.t.}\;\,\mathbf{0}\le\mathbf{C}\le\mathbf{\Upsilon}_{\mathbf{C}}, \nonumber
\end{align}
where $\mathbf{\hat{c}}_{i}$ denotes $i$-th row of $\mathbf{C}$. By following the previous discussion on $\mathbf{B}$, it can be shown that as long as vertices in $\mathcal{G}(\mathbf{A}^T\mathbf{A})$ are clustered, simultaneously optimizing eq.~\ref{ch3:eq32} -- \ref{ch3:eq34} leads to the item clustering indicator matrix $\mathbf{C}$.
\end{proof}


\subsection{A limitation of the NMF as a clustering method} \label{ch3:limitationnmf}

As shown in the proof of proposition \ref{ch3:prop1}, optimizing NMF objective is equivalent to applying the relaxed \emph{ratio association} to the item graph $\mathcal{G}(\mathbf{A}^T\mathbf{A})$ and the feature graph $\mathcal{G}(\mathbf{AA}^T)$ simultaneously. And because in the NMF, clustering membership of each point is directly determined by finding the largest projection on the axis of the decomposition rank subspace ($K$ subspace) \cite{Xu}, the NMF can only of\mbox{}fer good results if the data points are linearly separable. 

This is not the case with the spectral clustering, where the memberships are indirectly determined by applying k-means clustering on the resulting factors. This additional step can sometimes find correct assignments even though the data points are not linearly separable. And unfortunately, since the factors produced by the NMF are nonnegative and directly point to the cluster's centers \cite{Xu}, applying k-means clustering on the factors won't change the clustering assignments.

\begin{figure}[t]
 \begin{center}
  \subfigure[$\alpha=0.05$]{
   \includegraphics[width=0.22\textwidth]{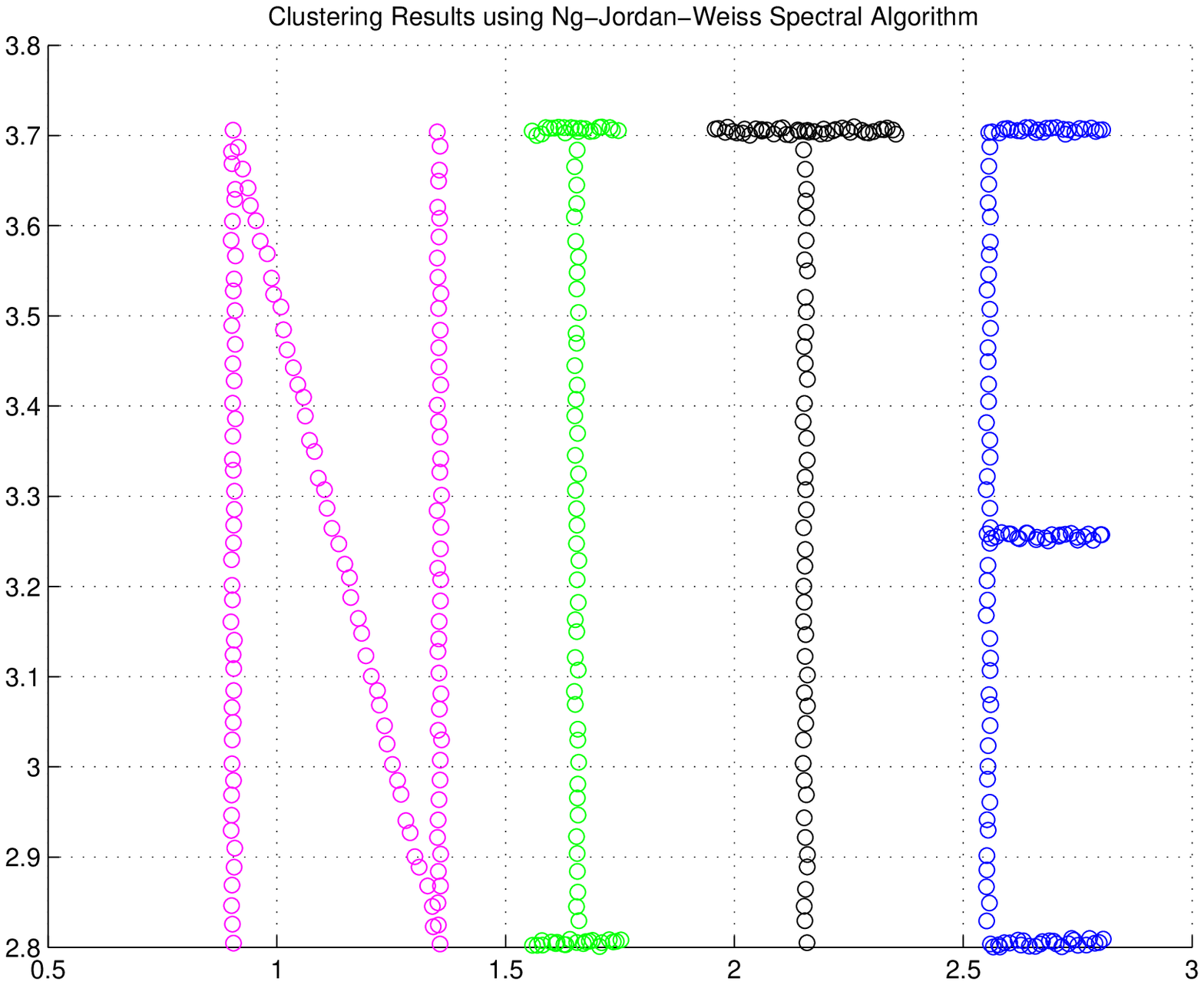}
   \label{ch2:fig1a}
  }
  \subfigure[$\alpha=0.1$]{
   \includegraphics[width=0.22\textwidth]{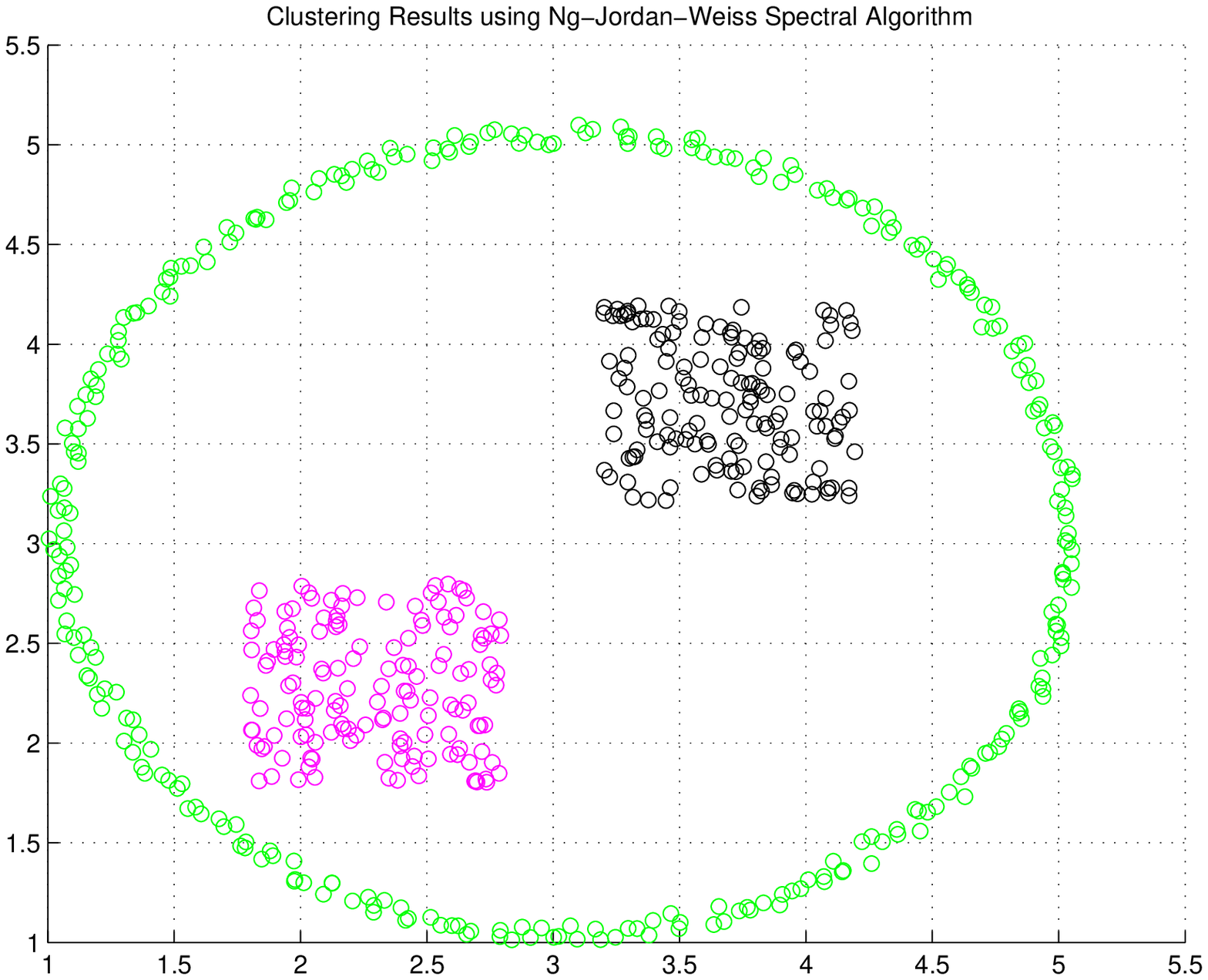}
   \label{ch2:fig1b}
  }
  \subfigure[$\alpha=0.1$]{
   \includegraphics[width=0.22\textwidth]{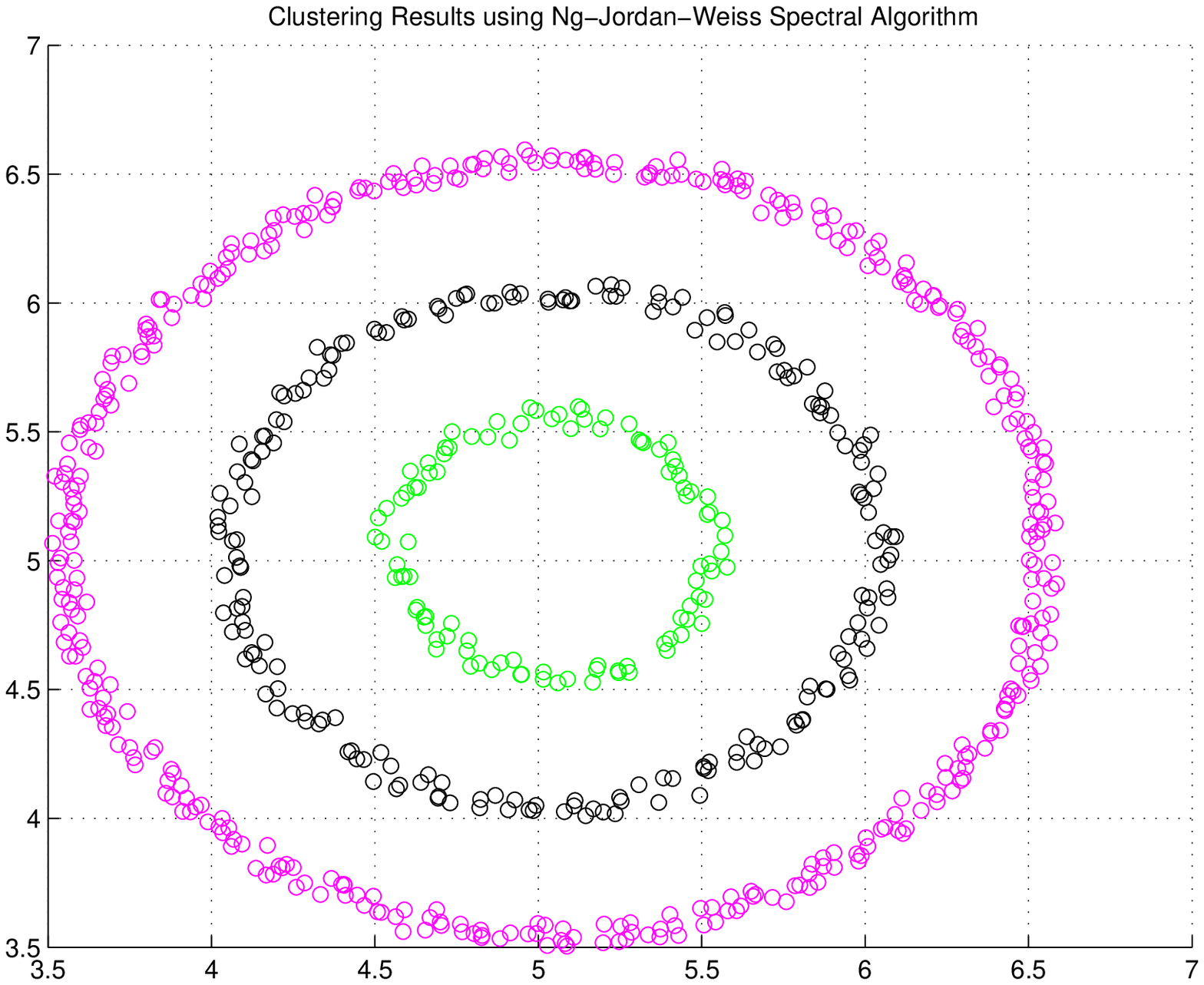}
   \label{ch2:fig1c}
  }
  \subfigure[$\alpha=0.2$]{
   \includegraphics[width=0.22\textwidth]{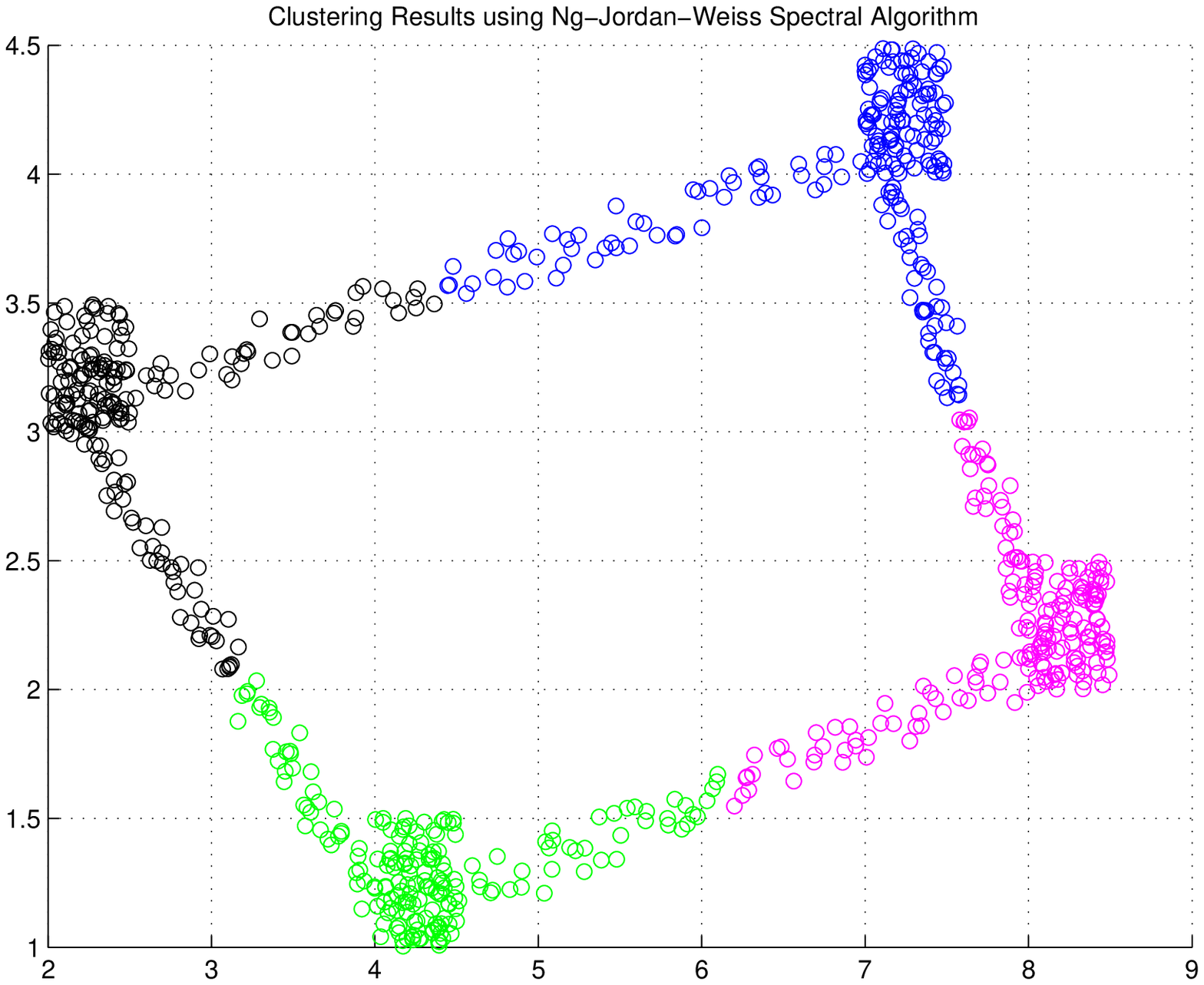}
   \label{ch2:fig1d}
  }
  \caption{Clustering linearly inseparable datasets using NJW.}
  \label{ch2:fig1}
 \end{center}
\end{figure}

\begin{figure}[t]
 \begin{center}
  \subfigure[$\alpha=0.3$]{
   \includegraphics[width=0.22\textwidth]{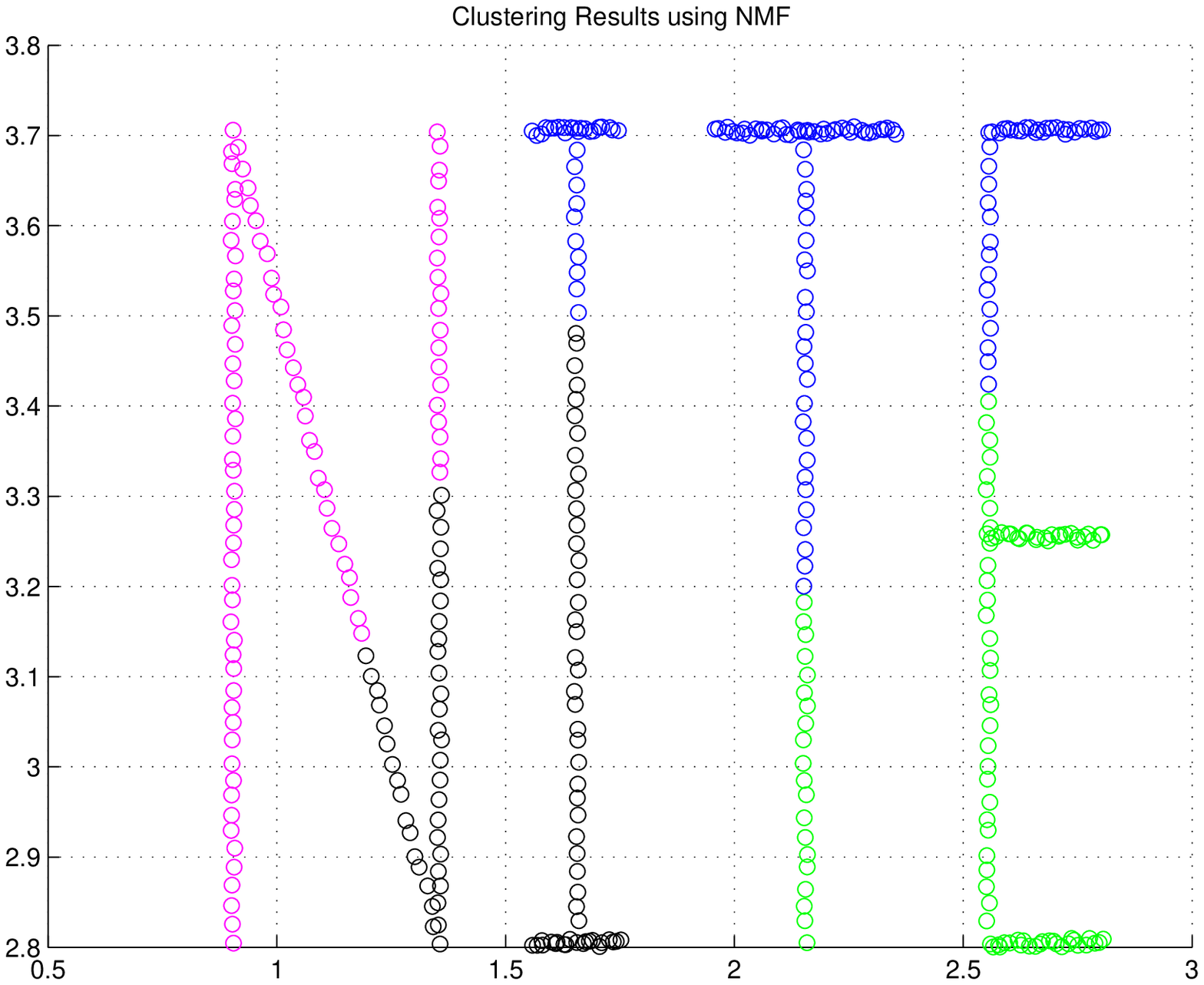}
   \label{ch3:fig1a}
  }
  \subfigure[$\alpha=0.5$]{
   \includegraphics[width=0.22\textwidth]{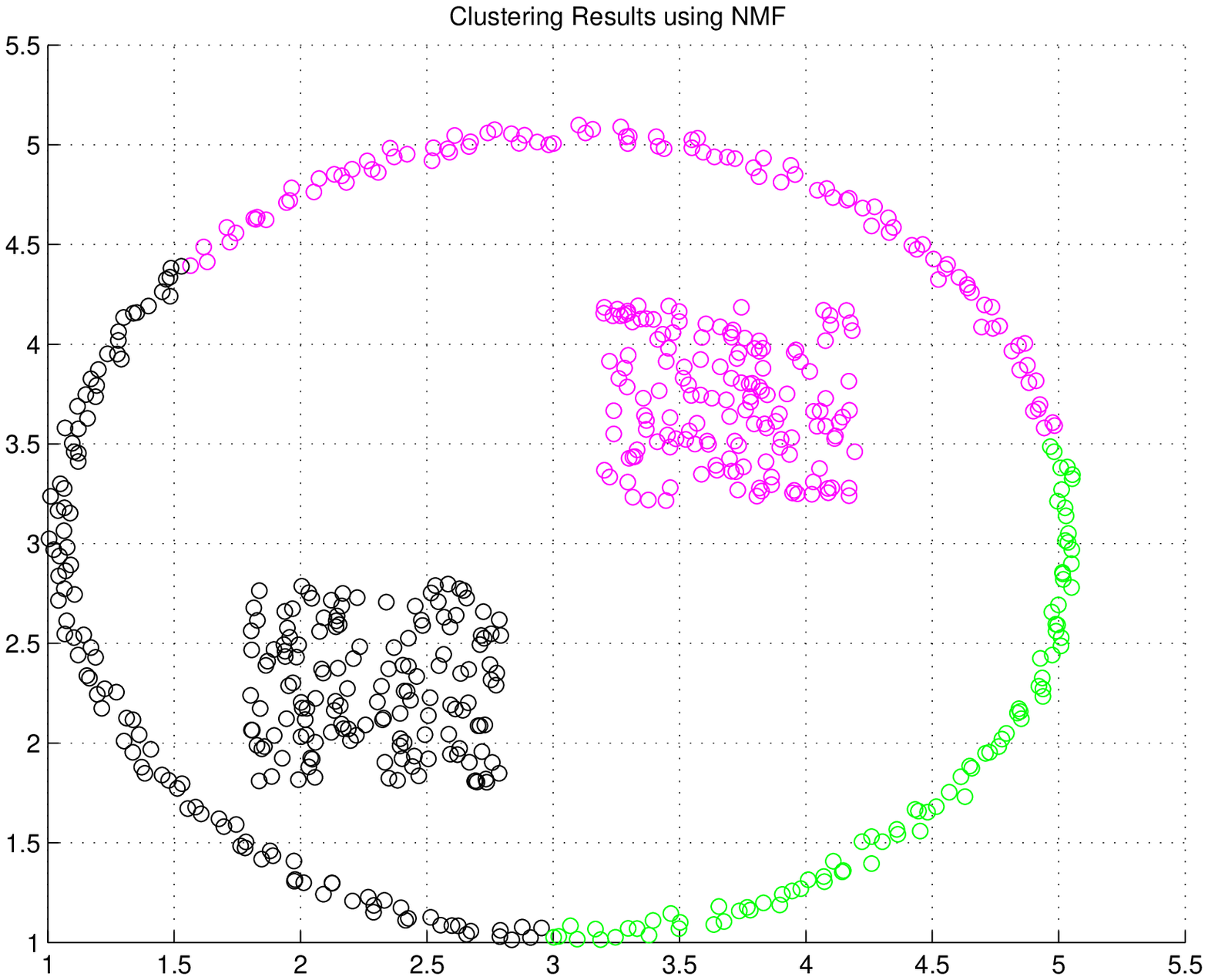}
   \label{ch3:fig1b}
  }
  \subfigure[$\alpha=0.4$]{
   \includegraphics[width=0.22\textwidth]{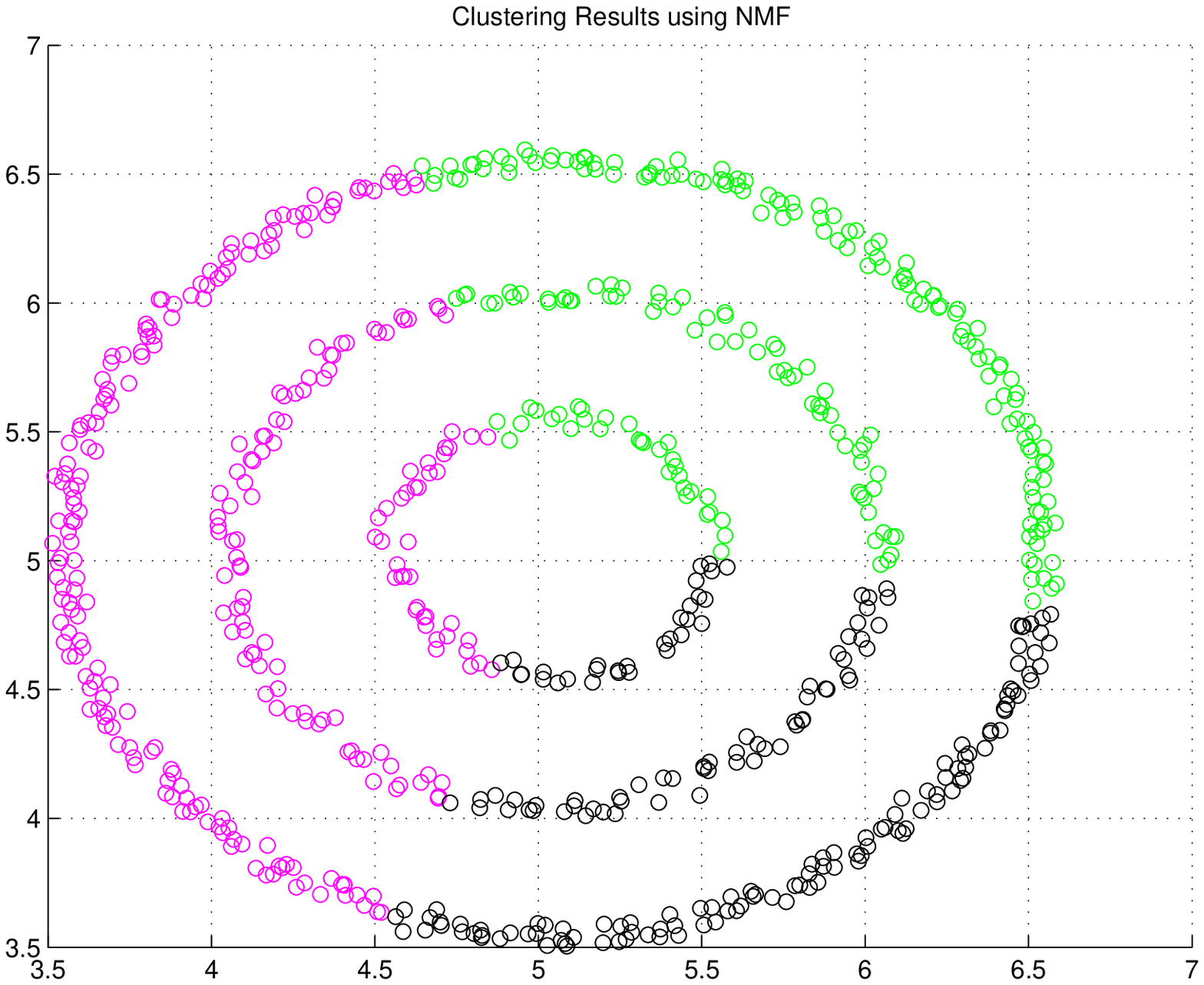}
   \label{ch3:fig1c}
  }
  \subfigure[$\alpha=0.5$]{
   \includegraphics[width=0.22\textwidth]{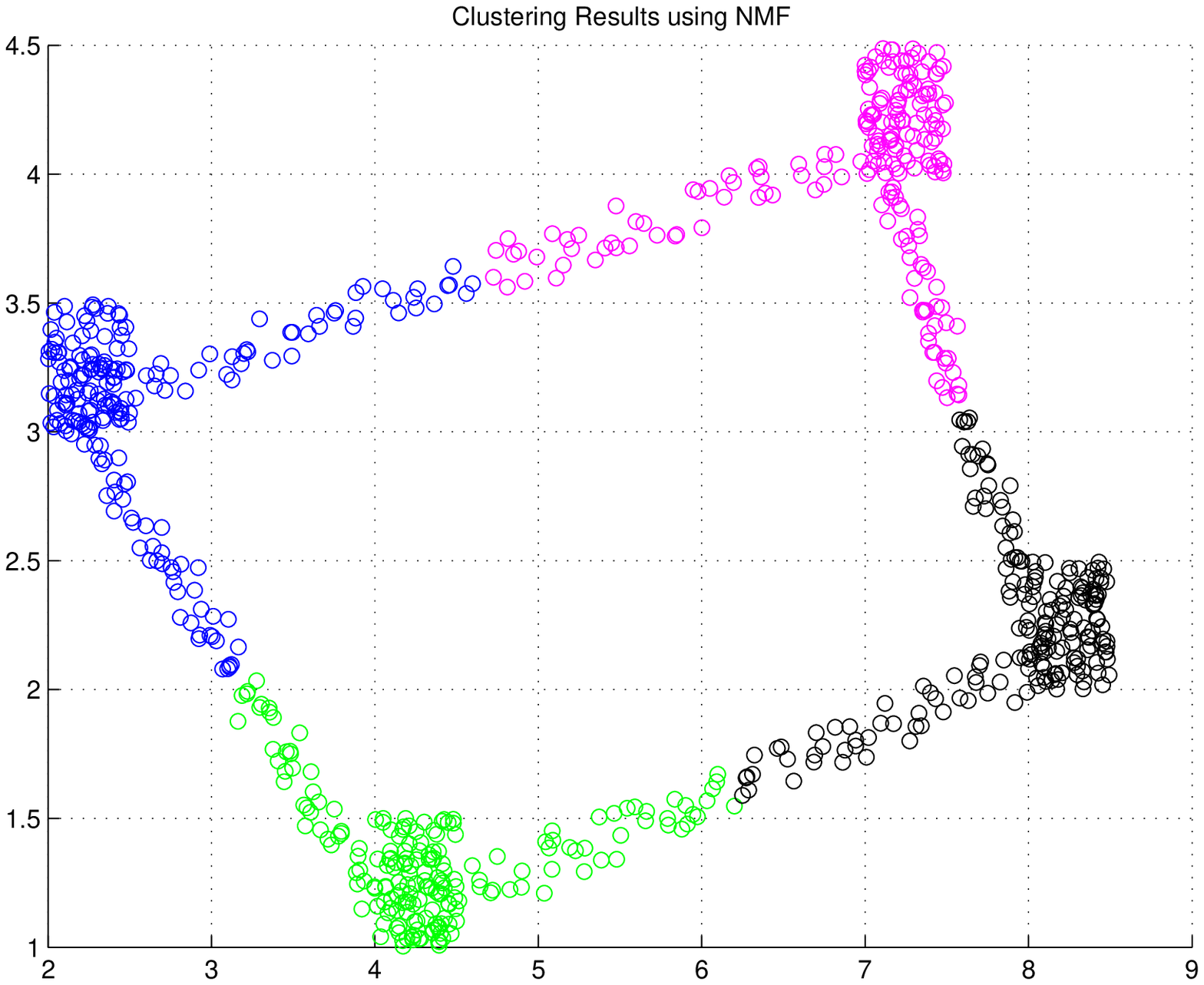}
   \label{ch3:fig1d}
  }
  \caption{Clustering linearly inseparable datasets using NMFLS.}
  \label{ch3:fig1}
 \end{center}
\end{figure}

\begin{figure}[t]
 \begin{center}
  \subfigure[$\alpha=0.8$]{
   \includegraphics[width=0.22\textwidth]{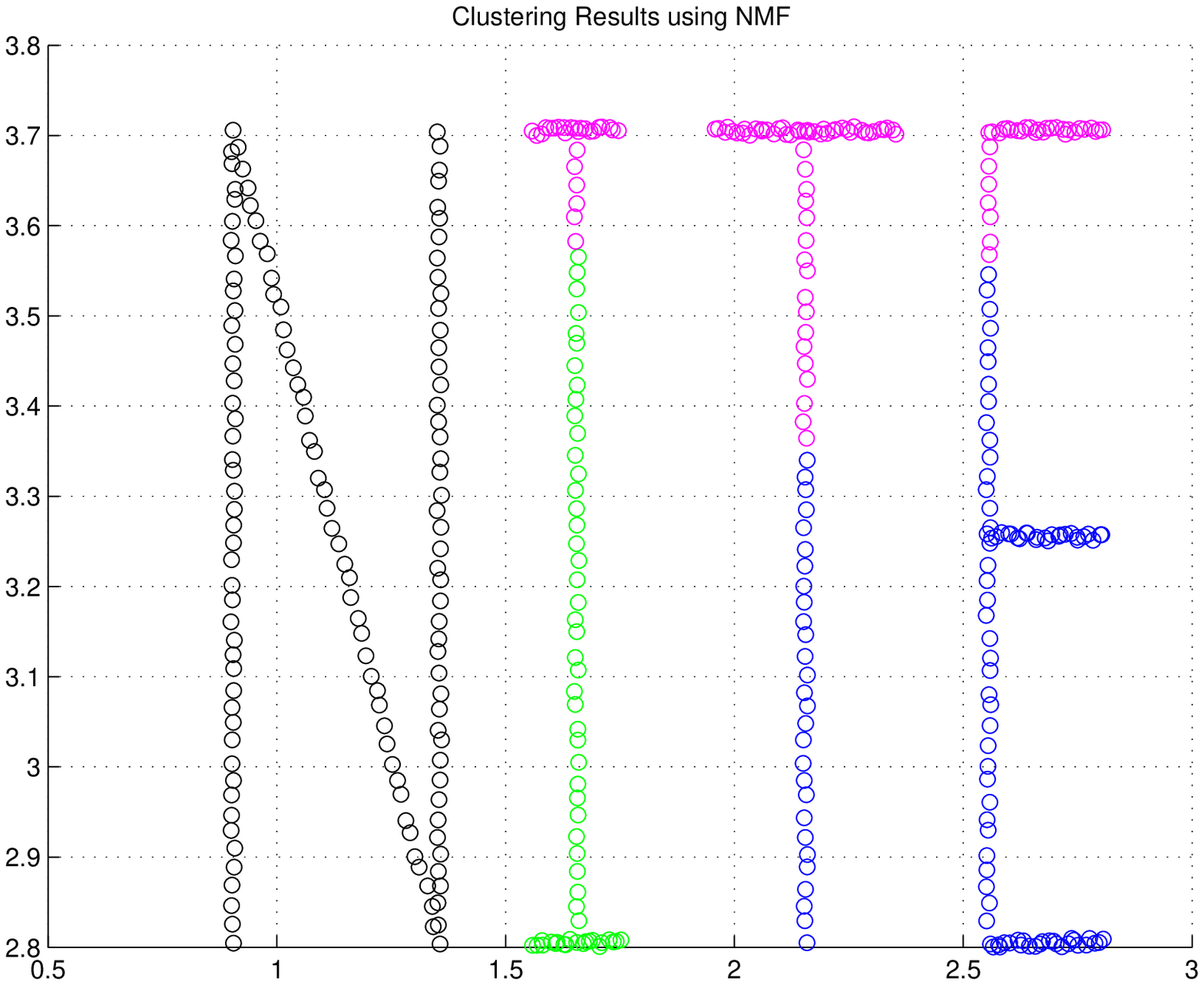}
   \label{ch3:fig2a}
  }
  \subfigure[$\alpha=0.6$]{
   \includegraphics[width=0.22\textwidth]{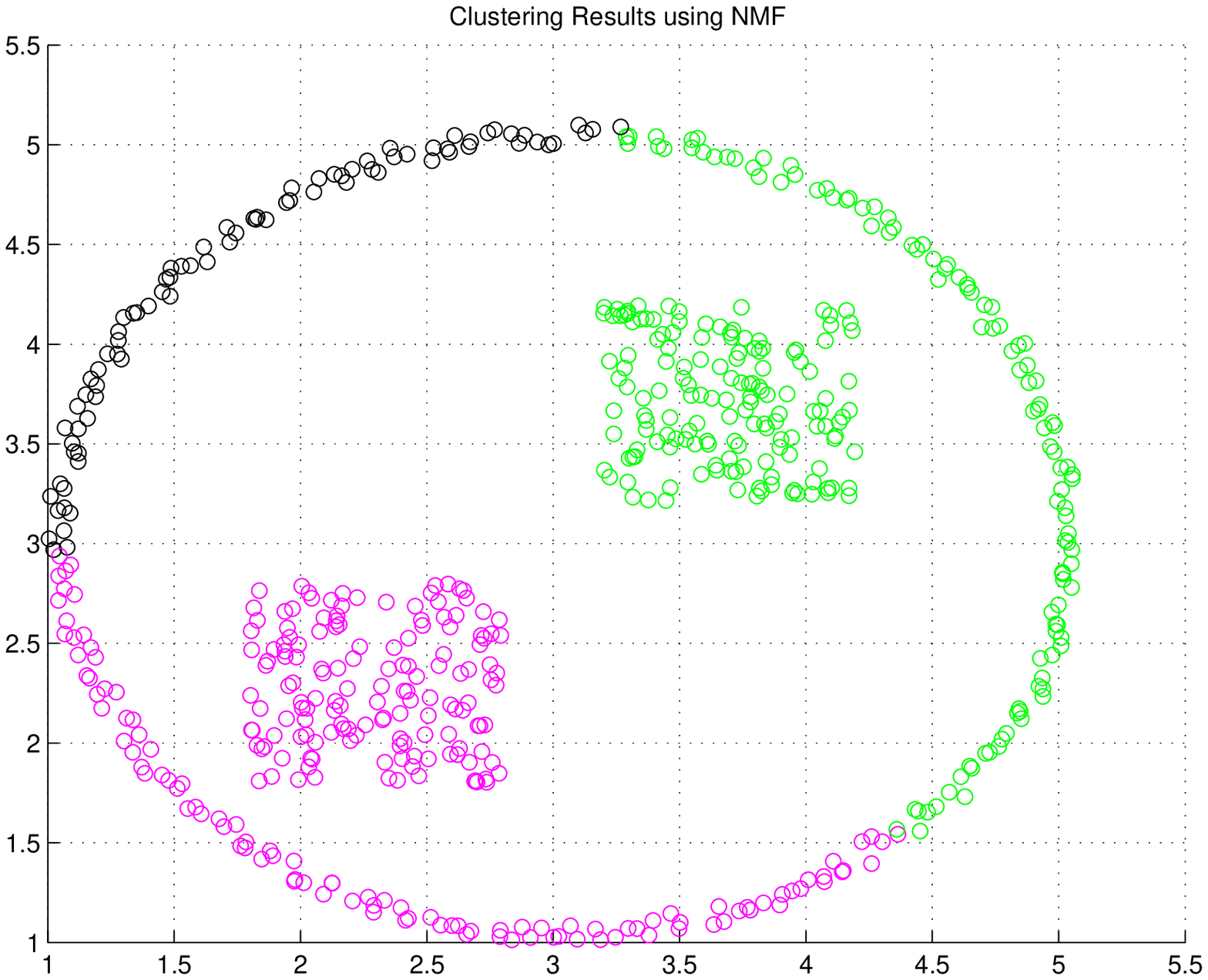}
   \label{ch3:fig2b}
  }
  \subfigure[$\alpha=0.4$]{
   \includegraphics[width=0.22\textwidth]{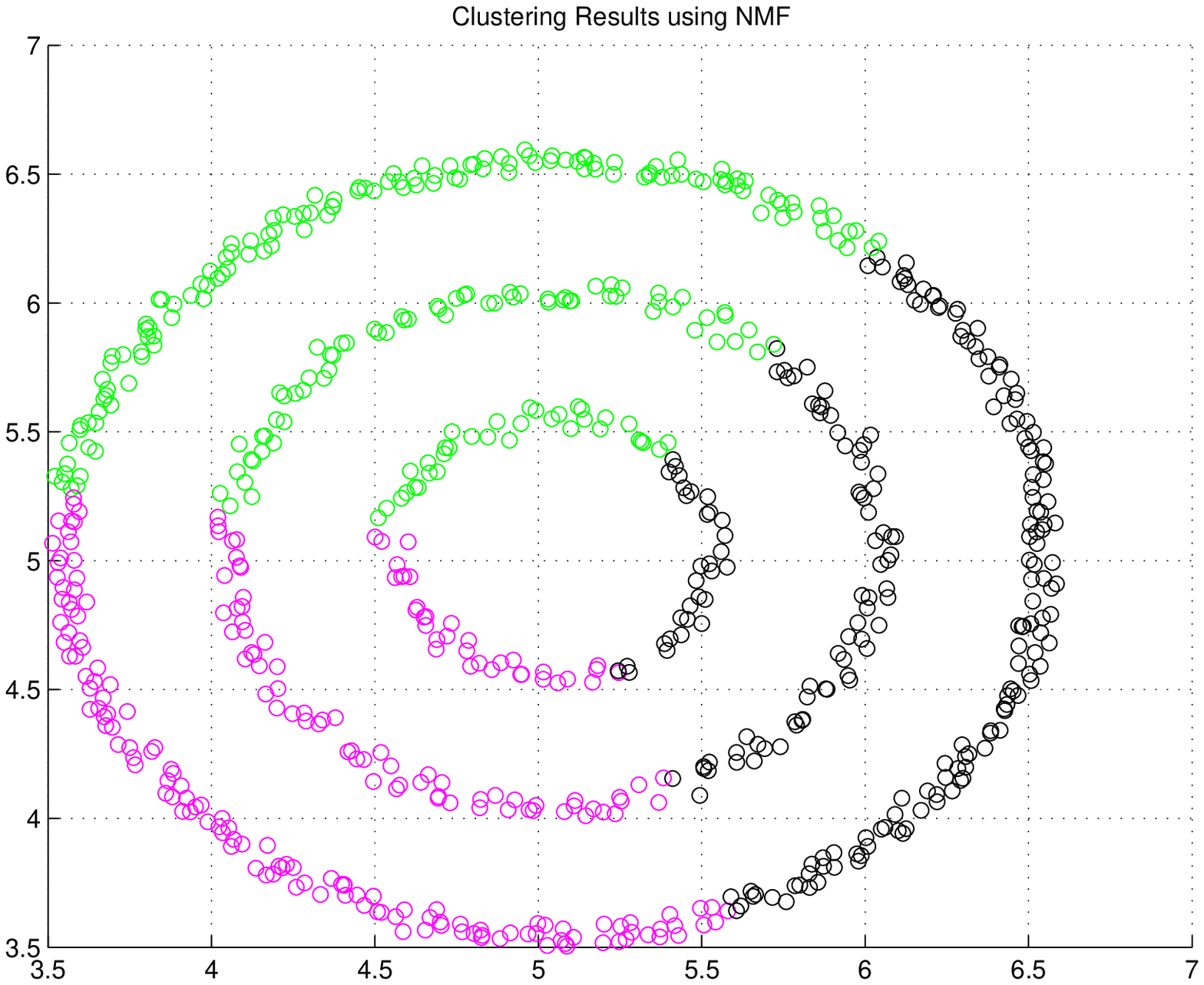}
   \label{ch3:fig2c}
  }
  \subfigure[$\alpha=0.4$]{
   \includegraphics[width=0.22\textwidth]{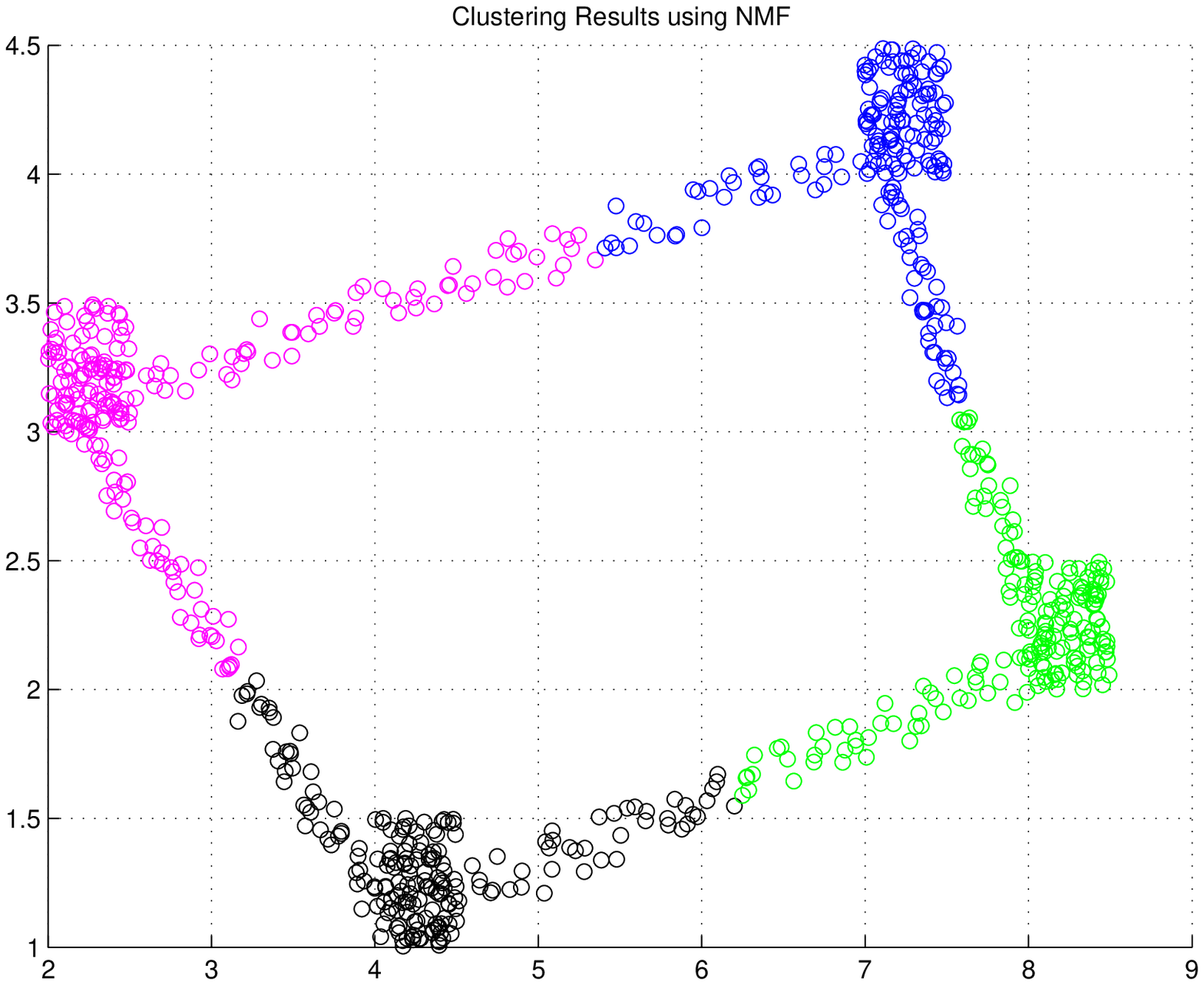}
   \label{ch3:fig2d}
  }
  \caption{Clustering linearly inseparable datasets using NMFJK.}
  \label{ch3:fig2}
 \end{center}
\end{figure}

The following examples show the limitation of the NMF in clustering linearly inseparable data points. And for comparison, the spectral clustering is used. For the spectral clustering, we use Ng et al.~algorithm (NJW) \cite{Ng}, and for the NMF, we use Lee \& Seung algorithm (NMFLS) \cite{Lee2}, and Kim \& Park algorithm (NMFJK) \cite{JKim2}. NJW and NMFLS are the standard algorithm for the spectral clustering and the NMF respectively, and NMFJK is the NMF algorithm that has convergence guarantee. Algorithm \ref{ch2:alg1} describes NJW algorithm, and algorithm \ref{ch3:alg2} describes clustering using the NMF. Note that we wrote codes for NJW and NMFLS by ourselves, and use codes from the authors website\protect\footnote{http://www.cc.gatech.edu/\~{}jingu/nmf/index.html} for NMFJK. To get the same treatment as in NJW, we use the same kernel strategy for NMFLS and NMFJK. The adjustable parameter $\alpha$ is learned directly from the datasets, and the results are displayed in figure \ref{ch2:fig1}, \ref{ch3:fig1}, and \ref{ch3:fig2}.

\begin{algorithm}
\caption{Spectral clustering algorithm by Ng et al.~\cite{Ng} (NJW).}
\label{ch2:alg1}
\begin{algorithmic}
\STATE \begin{enumerate}
\item Input: Rectangular data matrix $\mathbf{A}\in\mathbb{R}^{M\times N}$ with $N$ data points, \#cluster $K$, and Gaussian kernel parameter $\alpha$.
\item Construct symmetric af\mbox{}finity matrix $\mathbf{\dot{A}}\in\mathbb{R}^{N\times N}$ from $\mathbf{A}$ by using Gaussian kernel.
\item Normalize $\mathbf{\dot{A}}$ by $\mathbf{\dot{A}}\leftarrow\mathbf{D}^{-1/2}\mathbf{\dot{A}}\mathbf{D}^{-1/2}$ where $\mathbf{D}$ is a diagonal matrix with $D_{ii}=\sum_j\dot{a}_{ij}$.
\item Compute the $K$ largest eigenvectors of $\mathbf{\dot{A}}$, and form $\mathbf{\hat{X}}\in\mathbb{R}^{N\times K}=[\mathbf{\hat{x}}_1,\ldots,\mathbf{\hat{x}}_K]$, where $\mathbf{\hat{x}}_k$ is the $k$-th largest eigenvector of $\mathbf{\dot{A}}$.
\item Normalize every row of $\mathbf{\hat{X}}$, i.e., $X_{ij}\leftarrow X_{ij}/(\sum_j X_{ij}^2)^{1/2}$.
\item Apply k-means clustering on the row of $\mathbf{\hat{X}}$ to obtain the clustering indicator matrix $\mathbf{\bar{X}}\in\mathbb{R}^{N\times K}$.
\end{enumerate}
\end{algorithmic}
\end{algorithm}

\begin{algorithm}
\caption{Clustering by using the NMF.}
\label{ch3:alg2}
\begin{algorithmic}
\STATE \begin{enumerate}
\item Input: Rectangular data matrix $\mathbf{A}\in\mathbb{R}^{M\times N}$ with $N$ data points, \#cluster $K$, and Gaussian kernel parameter $\alpha$.
\item Construct symmetric af\mbox{}finity matrix $\mathbf{\dot{A}}\in\mathbb{R}^{N\times N}$ from $\mathbf{A}$ by using Gaussian kernel.
\item Compute $\mathbf{B}$ and $\mathbf{C}$ by using NMF algorithm (NMFLS or NMFJK) so that $\mathbf{\dot{A}}\approx\mathbf{BC}$.
\item Assume $\mathbf{C}$ is used, then clustering assignment of data point $n$, $x_n$, can be computed by $x_n\longleftarrow\nut_k\max\mathbf{c}_n,\;\forall n$.
\end{enumerate}
\end{algorithmic}
\end{algorithm}


As shown in figure \ref{ch2:fig1}, \ref{ch3:fig1}, and \ref{ch3:fig2}, while the spectral clustering can correctly find the clustering assignments for all datasets, the NMFs can only compete with the spectral clustering for the last dataset which is rather linearly separable. These results are in accord with the proof of proposition \ref{ch3:prop1} (that states as long as vertices on the feature $\langle$item$\rangle$ graph are clustered, optimizing the NMF objective leads to the feature $\langle$item$\rangle$ clustering indicator matrix). Thus, it seems that as a clustering method, the NMF is more similar to k-means clustering or support vector machine (SVM) which also can only cluster linearly separable datasets, than to the spectral methods, even though both clustering using the NMF and the spectral methods are based on matrix decomposition techniques. Accordingly, clustering performances of the NMF can probably be improved by using appropriate kernel methods as in k-means clustering and SVM.

\subsection{Experimental results} \label{ch3:results1}

The experiments are conducted to evaluate the performances of the NMF as a clustering method. All algorithms are developed in GNU Octave under linux platform using a notebook with 1.86 GHz Intel processor and 2 GB RAM. Reuters-21578 document corpus\footnote{http://kdd.ics.uci.edu/databases/reuters21578/reuters21578.html}, the standard dataset for testing learning algorithms and other text-based processing methods, is used for this purpose. This dataset contains 21578 documents (divided into 22 files with each file contains 1000 documents and the last file contains 578 documents) with 135 topics created manually with each document is assigned to one or more topics based on its content. The dataset is available in SGML and XML format, we use the XML version. We use all but the 18$^{\text{th}}$ file because this file is invalid both in its SGML and XML version. We use only documents that belong to exclusively one class (we use ``classes'' for refeering to the original grouping, and ``clusters'' for referring to groups resulted from the clustering algorithms). 

Further, we remove the common English stop words\protect\footnote{http://snowball.tartarus.org/algorithms/english/stop.txt}, stem the remaining words using Porter stemmer \cite{Rijsbergen}, and then remove words that belong to only one document. And also, we normalize the term-by document matrix $\mathbf{A}$ by: $\mathbf{A} \leftarrow \mathbf{AD}^{-1/2}$ where $\mathbf{D}=\text{diag}\big( \mathbf{A}^T\mathbf{A}\mathbf{e}\big)$ as suggested by Xu et al.~\cite{Xu}. We form test datasets by combining top 2, 4, 6, 8, 10, and 12 classes from the corpus. Table \ref{ch2:table3} summarizes the statistics of these test datasets, where \#doc, \#word, \%nnz, max, and min refer to the number of document, the number of word, percentage of nonzero entry, maximum cluster size, and minimum cluster size respectively. And table \ref{ch2:table4} gives the sizes (\#doc) of these top 12 classes.

\begin{table}
  \begin{center}
    \caption{Statistics of the test datasets.}
    \centering
    \begin{tabular}{|l|r|r|r|r|r|}
    \hline
    The data &\#doc & \#word & \%nnz & max & min \\
    \hline
    Reuters2    & 6090 & 8547  & 0.363 & 3874 & 2216 \\
    Reuters4    & 6797 & 9900  & 0.353 & 3874 & 333 \\
    Reuters6    & 7354 & 10319 & 0.347 & 3874 & 269 \\
    Reuters8    & 7644 & 10596 & 0.340 & 3874 & 144 \\
    Reuters10   & 7887 & 10930 & 0.336 & 3874 & 114 \\
    Reuters12   & 8052 & 11172 & 0.333 & 3874 &  75 \\
    \hline
    \end{tabular}
    \label{ch2:table3}
  \end{center}
\end{table}

\begin{table}
  \begin{center}
    \caption{Sizes of the top 12 classes.}
    \centering
    \begin{tabular}{|r|r|r|r|r|r|r|r|r|r|r|r|}
    \hline
    class &    1 &    2 &   3 &   4 &   5 &   6 \\ 
    \#doc & 3874 & 2216 & 374 & 333 & 288 & 269 \\ \hline
    class &   7 &   8 &   9 &  10 & 11 & 12 \\ 
    \#doc & 146 & 144 & 129 & 114 & 90 & 75 \\ \hline
    \end{tabular}
    \label{ch2:table4}
  \end{center}
\end{table}

As shown in table \ref{ch2:table3}, a rectangular word-by-document matrix $\mathbf{A}$, where $a_{ij}$ denotes the (weighted) frequency of word $i$ in document $j$, can be induced from each dataset (Reuters2, $\ldots$, Reuters12). Thus we have two options: either by directly applying co-clustering on bipartite graph $\mathcal{G}\big(\mathbf{A}\big)$ for simultaneously finding the word and document clustering, or by first transforming $\mathcal{G}\big(\mathbf{A}\big)$ into the corresponding unipartite graph $\mathcal{G}\big(\Phi(\mathbf{A}^T, \mathbf{A})\big)$ using kernel function $\Phi$, and then applying previously discussed clustering methods (algorithm \ref{ch2:alg1} and \ref{ch3:alg2}) for finding the document clustering. Employing kernel methods is unheard in document clustering researches (probably because of the sizes of the datasets), thus the co-clustering style will be employed instead. And actually, this is the most common way in using the NMF for clustering purpose \cite{Xu, Shahnaz, HKim2, Lee, Hoyer, Pauca, Ding2, SZLi, DKim, DKim2, HKim, JKim2, TLi, Ding0, Ding1, Cai, Gu}.

For comparison, we will employ the spectral co-clustering on $\mathcal{G}\big(\mathbf{A}\big)$ which is computed by finding the first $K$ singular vectors of $\mathbf{A}$. And because only reference classes for documents are available, we will only evaluate document clustering performances. The following theorem gives a theoretical support for the using of the SVD in \emph{multiclass spectral co-clustering} (more detailed discussion on this topic can be found in ref.~\cite{Mirzal2}), and algorithm \ref{ch2:alg2} and \ref{ch3:alg3} summarize the document clustering using the SVD and the NMF respectively.

\begin{theorem} \label{ch2:th6}
The optimal value of the following problem:
\begin{equation}
\max_{\mathbf{X}^T\mathbf{X}=\mathbf{Y}^T\mathbf{Y}=\mathbf{I}_K} \mathrm{tr}(\mathbf{X}^T\mathbf{R}\mathbf{Y}),
\label{ch2:eq14}
\end{equation}
is equal to $\sum_{k=1}^{K}\sigma_k$ if 
\begin{align*}
\mathbf{X} &= [\mathbf{x}_1,\ldots,\mathbf{x}_K]\mathbf{Q},\;\,\text{and}
\\
\mathbf{Y} &= [\mathbf{y}_1,\ldots,\mathbf{y}_K]\mathbf{Q}
\end{align*}
where $\mathbf{R}\in\mathbb{C}^{M\times N}$ denotes a full rank rectangular complex matrix with singular values $\sigma_1\ge\ldots\ge\sigma_{\min(M,N)}>0$, $0\le K\le\min(M,N)$, $\mathbf{X}\in\mathbb{C}^{M\times K}$ and $\mathbf{Y}\in\mathbb{C}^{N\times K}$ denote column orthogonal matrices, $\mathbf{x}_k$ and $\mathbf{y}_k$ ($k\in[1,K]$) respectively denote $k$-th left and right singular vectors  correspond to $\sigma_k$, and $\mathbf{Q}\in\mathbb{C}^{K\times K}$ denotes an arbitrary unitary matrix.
\end{theorem}

\begin{proof}
Eq.~\ref{ch2:eq14} can be rewritten as:
\begin{equation*}
\max_{\mathbf{X}^T\mathbf{X}=\mathbf{Y}^T\mathbf{Y}=\mathbf{I}_K}\frac{1}{2} \mathrm{tr}\left(\left[\begin{array}{c}\mathbf{X}\\\mathbf{Y}\end{array}\right]^T\underbrace{\left[\begin{array}{cc}\mathbf{0} & \mathbf{R}\\
\mathbf{R}^T & \mathbf{0}\end{array}\right]}_{\mathbf{\Psi}}\left[\begin{array}{c}\mathbf{X}\\\mathbf{Y}\end{array}\right]\right).
\end{equation*}
Since $\mathbf{\Psi}$ is a full rank Hermitian matrix, by the Ky Fan theorem (shown in theorem \ref{ch2:th2} below), the global optimum solution is given by the first $K$ eigenvectors of $\mathbf{\Psi}$:
\begin{equation*}
\left[\begin{array}{c}\mathbf{X}\\\mathbf{Y}\end{array}\right]=
\left[\begin{array}{c}\mathbf{x}_1,\ldots,\mathbf{x}_K\\
\mathbf{y}_1,\ldots,\mathbf{y}_K\end{array}\right]\mathbf{Q}.
\end{equation*}
Therefore, 
\begin{equation*}
\left[\begin{array}{cc}\mathbf{0} & \mathbf{R}\\
\mathbf{R}^T & \mathbf{0}\end{array}\right]\left[\begin{array}{c}\mathbf{x}_k\\
\mathbf{y}_k\end{array}\right] =
\lambda_k\left[\begin{array}{c}\mathbf{x}_k\\
\mathbf{y}_k\end{array}\right],
\end{equation*}
where $k\in[1,K]$ and $\lambda_k$ denotes $k$-th eigenvalue of $\mathbf{\Psi}$. Then,
\begin{align*}
\mathbf{R}\mathbf{y}_k &= \lambda_k\mathbf{x}_k,\,\; \text{and} 
\\
\mathbf{R}^T\mathbf{x}_k &= \lambda_k\mathbf{y}_k. 
\end{align*}
where $\mathbf{u}_k$ and $\mathbf{y}_k$ denote the left and right singular vectors associated with singular value $\lambda_k$($=\sigma_k$) of $\mathbf{R}$.
\end{proof}

\begin{theorem}[Ky Fan \cite{Nakic,Zha}] \label{ch2:th2}
The optimal value of the following problem:
\begin{equation*}
\max_{\mathbf{X}^T\mathbf{X}=\mathbf{I}_K} \mathrm{tr}(\mathbf{X}^T\mathbf{H}\mathbf{X})
\end{equation*}
is equal to $\sum_{k=1}^{K}\lambda_k$ if
\begin{equation*}
\mathbf{X} = [\mathbf{u}_1,\ldots,\mathbf{u}_K]\mathbf{Q},
\end{equation*}
where $\mathbf{H}\in\mathbb{C}^{N\times N}$ denotes a full rank Hermitian matrix with eigenvalues $\lambda_1\ge\ldots\ge\lambda_N\in\mathbb{R}$, $1\le K \le N$, $\mathbf{X}\in\mathbb{C}^{N\times K}$ denotes a column orthogonal matrix, $\mathbf{I}_K$ denotes a $K\times K$ identity matrix, $\mathbf{u}_k\in\mathbb{C}^{N}$ denotes $k$-th eigenvector corresponds to $\lambda_k$, and $\mathbf{Q}\in\mathbb{C}^{K\times K}$ denotes an arbitrary unitary matrix.
\end{theorem}

Note that, even though theoretically $\lambda_k$ can be chosen so that $\lambda_k=\sigma_k\;\forall k$, numerically $\lambda_k$ can be negative. And also, numerically $\mathbf{X}$ \& $\mathbf{Y}$ constructed using eigenvectors of $\mathbf{\Psi}$ can be different from using singular vectors of $\mathbf{R}$. Therefore one should always use the SVD for computing $\mathbf{X}$ and $\mathbf{Y}$. And for convenience, we assume $\mathbf{R}$ to be of full rank. The similar result can be derived for non full rank $\mathbf{R}$. 

Theorem \ref{ch2:th6} gives the theoretical support for directly applying graph cuts to the bipartite graph $\mathcal{G}\big(\mathbf{A}\big)$ to get simultaneous row and column clustering or also known as (multiclass) spectral co-clustering. The following gives the objective of the multiclass spectral co-clustering:
\begin{equation*}
\max_{\mathbf{\bar{X}}^T\mathbf{\bar{X}}=\mathbf{\bar{Y}}^T\mathbf{\bar{Y}}=\mathbf{I}_K} \mathrm{tr}(\mathbf{\bar{X}}^T\mathbf{A}\mathbf{\bar{Y}}),
\end{equation*}
where $\mathbf{\bar{X}}\in\mathbb{R}_{+}^{M\times K}$ and $\mathbf{\bar{Y}}\in\mathbb{R}_{+}^{N\times K}$ denote the row and column clustering indicator matrices respectively. By relaxing the nonnegativity constraints, $\mathbf{\bar{X}}$ and $\mathbf{\bar{Y}}$ can be found by computing the first $K$ left and right singular vectors of $\mathbf{A}$.

There are some standard metrics in evaluating clustering quality. The most commonly used metrics are \emph{mutual information}, \emph{entropy}, and \emph{purity}. We will use these metrics together with an additional metric, \emph{Fmeasure}. In the following, the definitions of these metrics are outlined.

\begin{algorithm}
\caption{Document clustering using the SVD.}
\label{ch2:alg2}
\begin{algorithmic}
\STATE \begin{enumerate}
\item Input: Rectangular word-by-document matrix $\mathbf{A}\in\mathbb{R}^{M\times N}$, and \#cluster $K$.
\item Normalize $\mathbf{A}$ by: $\mathbf{A} \leftarrow \mathbf{AD}^{-1/2}$ where $\mathbf{D}=\text{diag}( \mathbf{A}^T\mathbf{A}\mathbf{e})$.
\item Compute the first $K$ right singular vectors of $\mathbf{A}$, and form $\mathbf{V}\in\mathbb{R}^{N\times K}=[\mathbf{v}_1,\ldots,\mathbf{v}_K]$, where $\mathbf{v}_k$ is the $k$-th right singular vector of $\mathbf{A}$.
\item Apply k-means clustering on rows of $\mathbf{V}$ to obtain document clustering indicator matrix $\mathbf{\bar{V}}\in\mathbb{R}^{N\times K}$.
\end{enumerate}
\end{algorithmic}
\end{algorithm}

\begin{algorithm}
\caption{Document clustering using the NMF.}
\label{ch3:alg3}
\begin{algorithmic}
\STATE \begin{enumerate}
\item Input: Rectangular word-by-document matrix $\mathbf{A}\in\mathbb{R}^{M\times N}$, and \#cluster $K$.
\item Normalize $\mathbf{A}$ by: $\mathbf{A} \leftarrow \mathbf{AD}^{-1/2}$ where $\mathbf{D}=\text{diag}( \mathbf{A}^T\mathbf{A}\mathbf{e})$.
\item Compute $\mathbf{C}$ by using NMF algorithm (NMFLS or NMFJK) so that $\mathbf{A}\approx\mathbf{BC}$.
\item Compute clustering assignment of $n$-th document by: $x_n\equiv\nut_k\max\mathbf{c}_n,\;\forall n$.
\end{enumerate}
\end{algorithmic}
\end{algorithm}

\emph{Mutual information} (MI) measures dependency between the clusters produced by the algorithms and the reference classes. The higher the MI, the most related the clusters with the classes, and therefore the better the clustering will be. It is shown that MI is a superior measure than \emph{purity} and \emph{entropy} \cite{Strehl} because it is tolerant to the dif\mbox{}ference between \#cluster and \#class. MI is defined with the following formula:
\begin{equation*}
MI \equiv \sum_{r=1}^{R}\sum_{s=1}^{S}p(r,s)\log_2\left(\frac{p(r,s)}{p(r)p(s)}\right),
\end{equation*}
where $r$ and $s$ denote the $r$-th cluster and $s$-th class respectively, $p(r,s)$ denotes the joint probability distribution function of the clusters and the classes, and $p(r)$ and $p(s)$ denote the marginal probability distribution functions of the clusters and the classes respectively. Note that because of inconsistency in the formulation of normalized MI (a more commonly used metric) in the literatures, we use MI instead. Accordingly, MI's values are comparable only for the same dataset.

\emph{Entropy} addresses the composition of classes in a cluster. It measures uncertainty in the cluster, thus the lower the \emph{entropy}, the better the clustering will be. Unlike MI, if there is discrepancy between \#cluster and \#class, \emph{entropy} won't be very indicative about the the clustering quality. \emph{Entropy} is defined with the following:
\begin{equation*}
entropy \equiv \frac{1}{N\log_2S}\sum_{r=1}^{R}\sum_{s=1}^{S}c_{rs}\log_2\frac{c_{rs}}{c_r},
\end{equation*}
where $N$ is the number of samples (\#doc for document clustering), $c_{rs}$ denotes the number of samples in $r$-th cluster that belong to $s$-th class, and $c_r$ denotes the size of $r$-th cluster.

\emph{Purity} is the most commonly used metric. It measures the percentage of the dominant class in a cluster, so the high the better. As in \emph{entropy}, \emph{purity} is also sensitive to the discrepancy between \#cluster and \#class. \emph{Purity} is defined with:
\begin{equation*}
purity = \frac{1}{N}\sum_{r=1}^R\max_s c_{rs}.
\end{equation*}

And \emph{Fmeasure} combines two concept in IR: \emph{recall} and \emph{precision}. \emph{Recall} measures the proportion of the retrieved relevant documents to all relevant documents, and \emph{precision} measures the proportion of the retrieved relevant documents to all retrieved documents. In the context of assessing clustering quality, \emph{Fmeasure} is defined with \cite{Andrews}:
\begin{align*}
Fmeasure \equiv \frac{1}{R}\sum_{r=1}^R F_r,\;\;F_r = 2\,\frac{precision_r \times recall_r}{precision_r + recall_r},
\end{align*}
where $precision_r$ and $recall_r$ denote the \emph{precision} and \emph{recall} of $r$-th cluster.

Clustering results of the SVD and the NMFs are shown in table \ref{ch3:table1}--\ref{ch3:table4}. And time comparisons are given in table \ref{ch3:table5}, with times for the SVD are the sum of SVD computational times and the times for performing k-means to obtain clustering assignments, and times for the NMF are simply the times for performing the NMF on the data matrices. Because we use SVD built-in function that is written in C and highly optimized, the computational times of the SVD are not really comparable to the computational times of the NMF algorithms which are written in Matlab/Octave scripts.

As shown in table \ref{ch3:table1}--\ref{ch3:table4}, in general, NMFJK performs as good as the SVD for all the metrics with NMFJK tends to be better for datasets with smaller \#clusters and the SVD for datasets with bigger \#clusters. Unfortunately, NMFLS which is the most popular NMF algorithm seems to only be able to give moderate results. The convergence guarantee of NMFJK can probably have some role here as converged algorithms usually can approximate the original matrices better than algorithms without convergence guarantee \cite{CJLin2, JKim2}.

\begin{table}
 \begin{center}
   \caption{Average {\bf mutual information} over 10 trials.}
   \centering
   \begin{tabular}{|l|r|r|r|}
   \hline
Data & SVD & NMFLS & NMFJK \\
\hline
Reuters2 & $\mathbf{0.4951610991}$ & 0.4039195065 & 0.4825151487 \\ 
Reuters4 & 0.7345796343 & 0.6287861424 & $\mathbf{0.7482158671}$ \\ 
Reuters6 & 0.8367879438 & 0.7945867871 & $\mathbf{0.9763402437}$ \\
Reuters8 & $\mathbf{1.0342492298}$ & 0.9228548694 & 1.0110485952  \\
Reuters10 & $\mathbf{1.1754008483}$ & 1.0415397095 & 1.1588735544 \\
Reuters12 & 1.0812313058 & 1.1325663319 & $\mathbf{1.2069251441}$ \\ 
\hline
  \end{tabular}
  \label{ch3:table1}
 \end{center}
\end{table}

\begin{table}
 \begin{center}
   \caption{Average {\bf entropy} over 10 trials.}
   \centering
   \begin{tabular}{|l|r|r|r|}
   \hline
Data & SVD & NMFLS & NMFJK \\
\hline
Reuters2 & $\mathbf{0.4506918576}$ & 0.5419334502 & 0.463337808 \\ 
Reuters4 & 0.3491263843 & 0.4020231303 & $\mathbf{0.3423082679}$ \\ 
Reuters6 & 0.3675835184 & 0.3839091543 & $\mathbf{0.3135973194}$ \\
Reuters8 & $\mathbf{0.3185428072}$ & 0.3556742607 & 0.3262763521 \\
Reuters10 & $\mathbf{0.2957115633}$ & 0.3360077814 & 0.3006867745 \\
Reuters12 & 0.3338525186 & 0.3195329752 & $\mathbf{0.2987911092}$ \\ 
\hline
  \end{tabular}
  \label{ch3:table2}
 \end{center}
\end{table}

\begin{table}
 \begin{center}
   \caption{Average {\bf purity} over 10 trials.}
   \centering
   \begin{tabular}{|l|r|r|r|}
   \hline
Data & SVD & NMFLS & NMFJK \\
\hline
Reuters2 & 0.8623973727 & 0.821543514 & $\mathbf{0.8688505747}$ \\ 
Reuters4 & $\mathbf{0.8394880094}$ & 0.7941739003 & 0.8234515227 \\ 
Reuters6 & 0.68180582 & 0.7451047049 & $\mathbf{0.8042697852}$
 \\
Reuters8 & $\mathbf{0.8178963893}$ & 0.7490580848 & 0.7780612245 \\
Reuters10 & $\mathbf{0.786103715}$ & 0.7312032458 & 0.7769747686 \\
Reuters12 & 0.6838052658 & 0.7387729757 & $\mathbf{0.7663686041}$ \\ 
\hline
  \end{tabular}
  \label{ch3:table3}
 \end{center}
\end{table}

\begin{table}
 \begin{center}
   \caption{Average {\bf Fmeasure} over 10 trials.}
   \centering
   \begin{tabular}{|l|r|r|r|}
   \hline
Data & SVD & NMFLS & NMFJK \\
\hline
Reuters2 & 0.8595171797 & 0.8190358778 & $\mathbf{0.865279454}$ \\ 
Reuters4 & 0.6255202581 & 0.5615436865 & $\mathbf{0.6960413891}$ \\ 
Reuters6 & 0.6487551603 & 0.4622471694 & $\mathbf{0.6488871315}$ \\
Reuters8 & $\mathbf{0.5043941779}$ & 0.4040827621 & 0.4680952898 \\
Reuters10 & $\mathbf{0.516367264}$ & 0.3800132312 & 0.4842865587 \\
Reuters12 & $\mathbf{0.4437491506}$ & 0.3567059141 & 0.4333021978 \\ 
\hline 
  \end{tabular}
  \label{ch3:table4}
 \end{center}
\end{table}

\begin{table}
 \begin{center}
   \caption{Average computational times over 10 trials (second).}
   \centering
   \begin{tabular}{|l|r|r|r|}
   \hline
Data & SVD & NMFLS & NMFJK \\
\hline
Reuters2 & 4.675 & 77.27 & 65.45 \\ 
Reuters4 & 6.315 & 108.8 & 86.32 \\ 
Reuters6 & 14.01 & 134.0 & 105.1 \\
Reuters8 & 18.17 & 158.4 & 128.3 \\
Reuters10 & 19.98 & 834.7 & 452.2 \\
Reuters12 & 21.23 & 1249 & 775.8 \\ 
\hline 
  \end{tabular}
  \label{ch3:table5}
 \end{center}
\end{table}

The computational times of NMFJK seems to be promising as it is faster than NMFLS for all datasets. Note that since NMFLS and NMFJK are written in Matlab/Octave script, improving the computational performances of these algorithms is highly possible. And according to Albright et al.~\cite{Albright}, some highly optimized NMF algorithms can be faster than SVD algorithms.

\section{LSI aspect of the NMF} \label{ch3:lsinmf}

This section is the second part of this paper. Here, we will first describe LSI aspect of the NMF by showing its capability in solving synonymy and polysemy problems in some synthetic datasets given that the semantic structures allow the problems to be revealed, and then evaluate this aspect more extensively by comparing the results with results of the standard LSI method---the (truncated) SVD---using real datasets.

\subsection{Synonymy problems} \label{ch3:synonymy}

Synonyms are different words with similar or almost similar meaning, for example \{university, college, institute\}, \{female, girl, woman\}, and \{book, novel, biography\} each is a set of synonyms. For improving \emph{recall} and \emph{precision} of an IR system, it is expected that the system is able to recognize the synonyms. This task is usually done by approximating the original word-by-document matrix with its truncated SVD version \cite{Deerwester, Berry2}. 

The synthetic dataset in table \ref{ch3:table6} (taken from ref.~\cite{Kolda}) shows the synonymy problems in which Mark Twain \& Samuel Clemens refer to the same person, and purple \& colour are closely related. As shown, Mark Twain \& Samuel Clemens are not recognized as the same person; and similarly, purple \& colour also are not recognized to be related. Accordingly, if a query $\mathbf{q}$ containing \{mark, twain\} ($\mathbf{q}^T=[1\;1\;0\;0\;0\;0]$) is made into the original matrix $\mathbf{A}$, then $\mathbf{q}^T\mathbf{A}=[30,0,20,0,0]$. So, only Doc1 and Doc3 are retrieved, and Doc2 is lost. Similarly, if a query containing \{colour\} is made, then only Doc4 will be retrieved, and Doc5 will be lost (note that even though purple and colour can have different meanings, according to this example they are highly related, thus any query containing either one of them is expected to retrieve Doc4 and Doc5).

\begin{table}
 \begin{center}
   \caption{Dataset for describing synonymy problems.}
   \centering
   \begin{tabular}{|l|r|r|r|r|r|}
   \hline
Word & Doc1 & Doc2 & Doc3 & Doc4 & Doc5 \\
\hline
mark & 15 & 0 & 0 & 0 & 0 \\ 
twain & 15 & 0 & 20 & 0 & 0 \\ 
samuel & 0 & 10 & 5 & 0 & 0 \\
clemens & 0 & 20 & 10 & 0 & 0 \\
purple & 0 & 0 & 0 & 20 & 10 \\
colour & 0 & 0 & 0 & 15 & 0 \\ 
\hline 
  \end{tabular}
  \label{ch3:table6}
 \end{center}
\end{table}

\begin{table}
 \begin{center}
   \caption{LSI using the SVD for detecting synonyms.}
   \centering
   \begin{tabular}{|l|r|r|r|r|r|}
   \hline
Word & Doc1 & Doc2 & Doc3 & Doc4 & Doc5 \\
\hline
mark & 3.7 & 3.5 & 5.5 & -$\epsilon$ & -$\epsilon$ \\ 
twain & 11 & 10 & 16 & -$\epsilon$ & -$\epsilon$ \\ 
samuel & 4.1 & 3.9 & 6.1 & -$\epsilon$ & -$\epsilon$ \\
clemens & 8.3 & 7.8 & 12 & -$\epsilon$ & -$\epsilon$ \\
purple & -$\epsilon$ & -$\epsilon$ & -$\epsilon$ & 21 & 7.1 \\
colour & -$\epsilon$ & -$\epsilon$ & -$\epsilon$ & 13 & 4.5 \\ 
\hline 
  \end{tabular}
  \label{ch2:table10}
 \end{center}
\end{table}

\begin{table}
 \begin{center}
   \caption{LSI using the NMF (NMFLS) for detecting synonyms.}
   \centering
   \begin{tabular}{|l|r|r|r|r|r|}
   \hline
Word & Doc1 & Doc2 & Doc3 & Doc4 & Doc5 \\
\hline
mark & 3.72 & 3.50 & 5.45 & 0    & 0    \\
twain & 11.0 & 10.4 & 16.2 & 0    & 0    \\
samuel & 4.15 & 3.90 & 6.08 & 0    & 0    \\
clemens & 8.29 & 7.79 & 12.1 & 0    & 0    \\
purple & 0    & 0    & 0    & 21.0 & 7.08 \\
colour & 0    & 0    & 0    & 13.5 & 4.55 \\ 
\hline 
  \end{tabular}
  \label{ch3:table7a}
 \end{center}
\end{table}

\begin{table}
 \begin{center}
   \caption{LSI using the NMF (NMFJK) for for detecting synonyms.}
   \centering
   \begin{tabular}{|l|r|r|r|r|r|}
   \hline
Word & Doc1 & Doc2 & Doc3 & Doc4 & Doc5 \\
\hline
mark & 3.14 & 2.95 & 4.60 & 0    & 0    \\
twain & 9.27 & 8.71 & 13.6 & 0    & 0    \\
samuel & 3.50 & 3.29 & 5.13 & 0    & 0    \\
clemens & 7.00 & 6.58 & 10.3 & 0    & 0    \\
purple & 0    & 0    & 0    & 17.3 & 5.83 \\
colour & 0    & 0    & 0    & 11.1 & 3.74 \\
\hline 
  \end{tabular}
  \label{ch3:table7b}
 \end{center}
\end{table}

The synonymy problem can be resolved using LSI technique as long as there is a path that chains them together given that the path is close enough \cite{Kontostathis}. For example in table \ref{ch3:table6} mark \& twain are connected to samuel \& clemens through Doc3. So, there is a path that connects them, and it happens that the distance is close. Thus we can expect that LSI using the SVD will be able to reveal this hidden relationship. Similarly, colour \& purple are connected through Doc4, thus LSI is also expected to be able to reveal this relationship. Table \ref{ch2:table10} shows the result of rank-2 matrix approximation by using the SVD to the original matrix in table \ref{ch3:table6} with $\epsilon$ denotes small positive number. Note that the rank is chosen based on the number of reference classes, i.e., author names (\{mark, twain, samuel, clemens\}) and colour related terms (\{purple, colour\}). As shown, Doc1, Doc2, and Doc3 are now indexing mark, twain, samuel, and clemens. Thus, any query containing any of these words will correctly retrieve the corresponding relevant documents. And similarly, Doc4 and Doc5 are now indexing purple and colour, so any query containing at least one of these words will correctly retrieve the corresponding relevant documents. 

Now, we will apply the NMF to the data matrix in table \ref{ch3:table6} and see whether this technique can solve the synonymy problems. Table \ref{ch3:table7a} and \ref{ch3:table7b} show rank-2 matrix approximations using NMFLS and NMFJK respectively. As shown, both algorithms correctly index the synonyms, and thus the NMF can also be used in solving the synonymy problems in this dataset.

\subsection{Polysemy problems} \label{ch3:polysemy}

LSI technique is also expected to be able to solve polysemy---word with multiple unrelated meanings---problem. By using a synthetic dataset, we will describe how the standard LSI method and the NMF solve this problem, given that polyseme presents in unrelated documents. Table \ref{ch2:table11} gives an example of polyseme where bank can either refers to financial institution or area near river. By inspection, it is clear that the dataset contains two different topics: financial and river, with \{Doc1, Doc3, Doc5\} \& \{money, bank, interest\} are in the first topic; and \{Doc2, Doc4, Doc6\} \& \{bed, river, bank\} are in the second topic. Note that the dataset is well-conditioned for describing the polysemy problem as bank presents in unrelated documents. 

If a query $\mathbf{q}_1^T=[1\;0\;0\;1\;0]$ containing \{money, bank\} is made to the original matrix $\mathbf{A}$ in table \ref{ch2:table11}, then $\mathbf{q}_1^T\mathbf{A}=[2\;1\;2\;1\;1\;1]$. So, only Doc1 and Doc3 are recognized as relevant, and Doc5 will not be recognized as relevant. Similarly, if a query $\mathbf{q}_2^T=[0\;0\;1\;1\;0]$ containing \{river, bank\} is made, then $\mathbf{q}_2^T\mathbf{A}=[1\;2\;1\;2\;1\;1]$; only Doc2 and Doc4 are recognized as relevant, and Doc6 is not.

Table \ref{ch2:table12} shows rank-2 SVD approximation to the original matrix. If the same queries are made to the matrix $\mathbf{\hat{A}}$ in table \ref{ch2:table12}, then $\mathbf{q}_1^T\mathbf{\hat{A}}$ $=$ $[1.86726\;$ $1.00346\;$ $1.86726\;$ $1.00346\;$ $1.40251\;$ $0.91744]$ and $\mathbf{q}_2^T\mathbf{\hat{A}}$ $=$ $[1.00346\;$ $1.86726\;$ $1.00346\;$ $1.86726\;$ $0.91744\;$ $1.40251\;]$. Therefore, all relevant documents can be correctly retrieved, so LSI using the SVD can solve the polysemy problem in this dataset.

\begin{table}[!ht]
 \begin{center}
   \caption{Dataset for describing polysemy problem.}
   \centering
   \begin{tabular}{|l|r|r|r|r|r|r|}
   \hline
Word & Doc1 & Doc2 & Doc3 & Doc4 & Doc5 & Doc6 \\
\hline
money    & 1 & 0 & 1 & 0 & 0 & 0 \\ 
bed      & 0 & 1 & 0 & 1 & 0 & 1 \\ 
river    & 0 & 1 & 0 & 1 & 0 & 0 \\
bank     & 1 & 1 & 1 & 1 & 1 & 1 \\
interest & 1 & 0 & 1 & 0 & 1 & 0 \\
\hline 
  \end{tabular}
  \label{ch2:table11}
 \end{center}
\end{table}

\begin{table}[!ht]
 \begin{center}
   \caption{LSI using the SVD for detecting polyseme.}
   \centering
   \begin{tabular}{|l|r|r|r|r|r|r|}
   \hline
Word & Doc1 & Doc2 & Doc3 & Doc4 & Doc5 & Doc6 \\
\hline
money    & 0.80882 & -0.054983 & 0.80882 & -0.054983 & 0.547139 & 0.062068 \\ 
bed      & -0.023949 & 1.08239 & -0.023949 & 1.08239 & 0.117052 & 0.738319 \\ 
river    & -0.054983& 0.80882 & -0.054983 & 0.80882 & 0.062068 & 0.547139 \\
bank     & 1.05844 & 1.05844 & 1.05844 & 1.05844 & 0.855371 & 0.855371 \\
interest & 1.08239 & -0.023949 & 1.08239 & -0.023949 & 0.738319 & 0.117052 \\
\hline 
  \end{tabular}
  \label{ch2:table12}
 \end{center}
\end{table}

\begin{table}[!ht]
 \begin{center}
   \caption{LSI using the NMF (NMFLS) for detecting polyseme.}
   \centering
   \begin{tabular}{|l|r|r|r|r|r|r|}
   \hline
Word & Doc1 & Doc2 & Doc3 & Doc4 & Doc5 & Doc6 \\
\hline
money    & 0.801054 & 0.013549 & 0.800924 & 0.013018 & 0.558518 & 0.082122  \\ 
bed      & 0.013619 & 1.082496 & 0.014032 & 1.082806 & 0.098272 & 0.748645  \\ 
river    & 0.010063 & 0.804439 & 0.01037 & 0.80467 & 0.072989 & 0.556338  \\
bank     & 1.067149 & 1.063328 & 1.067377 & 1.062928 & 0.829791 & 0.831112  \\
interest & 1.080788 & 0.018281 & 1.080612 & 0.017564 & 0.753557 & 0.110799  \\
\hline 
  \end{tabular}
  \label{ch3:table8}
 \end{center}
\end{table}

\begin{table}[!ht]
 \begin{center}
   \caption{LSI using the NMF (NMFJK) for detecting polyseme.}
   \centering
   \begin{tabular}{|l|r|r|r|r|r|r|}
   \hline
Word & Doc1 & Doc2 & Doc3 & Doc4 & Doc5 & Doc6 \\
\hline
money    & 0.63733 & 0.040462 & 0.63733 & 0.040462 & 0.441459 & 0.098975  \\ 
bed      & 0.054469 & 0.858063 & 0.054469 & 0.858063 & 0.133246 & 0.594351  \\ 
river    & 0.040458 & 0.637333 & 0.040458 & 0.637333 & 0.09897  & 0.441459  \\
bank     & 0.877087 & 0.877088 & 0.877087 & 0.877088 & 0.699338 & 0.699339  \\
interest & 0.85806 & 0.054476 & 0.85806 & 0.054476 & 0.594352 & 0.133253  \\
\hline 
  \end{tabular}
  \label{ch3:table9}
 \end{center}
\end{table}

Now, we will see whether the NMF can also solve the problem. Table \ref{ch3:table8} and \ref{ch3:table9} show rank-2 NMF approximations using NMFLS and NMFJK respectively. Let $\mathbf{A}_1$ and $\mathbf{A}_2$ be the matrix in table \ref{ch3:table8} and \ref{ch3:table9} respectively. If the same queries are made to these matrices, then $\mathbf{q}_1^T\mathbf{A}_1$ $=$ $[1.8682\;$ $1.07688\;$ $1.8683\;$ $1.07595\;$ $1.38831\;$ $0.91323]$, $\mathbf{q}_1^T\mathbf{A}_2$ $=$ $[1.51442\;$ $0.91755\;$ $1.51442\;$ $0.91755\;$ $1.1408\;$ $0.79831]$, $\mathbf{q}_2^T\mathbf{A}_1$ $=$ $[1.07721\;$ $1.86777\;$ $1.07775\;$ $1.8676\;$ $0.90278\;$ $1.38745]$, and $\mathbf{q}_2^T\mathbf{A}_2$ $=$ $[0.91754\;$ $1.51442\;$ $0.91754\;$ $1.51442\;$ $0.79831\;$ $1.1408]$. Accordingly, the polysemy problem can also be solved by using the NMF in this dataset.

\subsection{Experimental results} \label{ch3:results2}

\begin{table}[t]
 \begin{center}
   \caption{The standard text collections in LSI.}
   \centering
   \begin{tabular}{|l|r|r|r|r|}
   \hline
            & Medline & Cranfield & CISI    & ADI    \\
\hline
\#Doc & 1033    & 1398      & 1460    & 82     \\ 
\#Word     & 12011   & 6551      & 9080    & 1215   \\ 
\%NNZ       & 0.4567  & 0.85674   & 0.51701 & 2.1479 \\
\#Query   & 30      & 225       & 35      &  35    \\
\hline 
  \end{tabular}
  \label{ch2:table16}
 \end{center}
\end{table}

We will now evaluate LSI aspect of the NMF by using real datasets, and compare the results with the results of the SVD. Table \ref{ch2:table16} summarizes the datasets\protect\footnote{http://web.eecs.utk.edu/research/lsi/} used in the experiments where \#Doc, \#Word, \%NNZ, and \#Query denote the number of documents, the number of unique words, percentage of nonnegative entries, and the number of predefined queries made to the corresponding word-by-document matrix respectively. These datasets are the standard text collections which have been extensively used in the LSI researches.

Each of the text collections comprises of three important files. The first file contains abstracts of the documents which each indexed by a unique identifier, the second file contains the list of queries each with a unique identifier, and the third file contains a dictionary that maps every query with its manually assigned relevant documents. 

The first file is the file that is used to construct the word-by-document matrix $\mathbf{A}\in\mathbb{R}_+^{M\times N}$. To extract the unique words, the stop words and words that shorter than two characters are removed. But we do not employ any stemming and do not remove words that only belong to one documents as in section \ref{ch3:results1}. The reasons are the stemming process seems to be not popular in the LSI researches, and removing unique words in a document can potentially reduce \emph{recall} since it is possible that queries contain these words. Then after $\mathbf{A}$ is constructed, we further adjust the entry weights by using logarithmic scale, i.e., $A_{ij}\leftarrow \log(A_{ij}+1)$, but do not normalized the columns of the matrix. This is because based on our pre-experimental results, the logarithmic scale performs better than the simple frequency of word occurences, and normalization has a negative ef\mbox{}fect on the retrieval performances for both the SVD and the NMF for all text collections.  

The second file is used to construct the query matrix $\mathbf{Q}$ $\in$ $\mathbb{R}_+^{Q\times M}$ $=$ $[\mathbf{q}_1,$ $\ldots,$ $\mathbf{q}_Q$ $]^T$ where $Q$ denotes the number of queries (shown in the last row of table \ref{ch2:table16}), $M$ denotes the number of unique words, and $\mathbf{q}_q$ denotes the $q$-th query vector constructed from the file. So simply by multiplying $\mathbf{Q}$ with the corresponding $\mathbf{A}$, one can get a matrix that contains scores that describe how relevant each query to the documents in the corresponding row.

And the third file is the file that maps each query to its manually assigned relevant documents. This information will be utilized as the references to measure the retrieval performances of the SVD and the NMF.

To measure the LSI performances, \emph{average precision}---the standard metric in the IR researches \cite{Harman} that measures $I$-point interpolated average \emph{pseudo-precision} at \emph{recall} level $[0,1]$---will be used. This metric captures both \emph{recall} and \emph{precision} concepts without inheriting the weakness from \emph{recall}, i.e., perfect \emph{recall} can be achieved by retrieving all documents. The following outlines the \emph{average precision} definition, and more detailed discussions can be found in, e.g., ref.~\cite{Kolda, Harman, Berry3}.

First the definition of \emph{precision} will be discussed. Let $\mathbf{r}=\mathbf{q}^T\mathbf{A}$ denotes a vector that contains document scores with respect to the query vector $\mathbf{q}$, and let $\mathbf{r}$ be sorted in reverse order (larger comes first). The precision at $n$-th document is given by:
\begin{equation*}
p_n \equiv \frac{r_n}{n}.
\end{equation*}
where $r_n$ denotes the number of relevant documents up to $n$-th position. The \emph{pseudo-precision} at recall level $x\in[0,1]$ is defined as:
\begin{equation*}
\hat{p}(x) \equiv \max\{p_n\;|\;x\le r_n/r_N,\;\,n=1,\ldots,N\},
\end{equation*}
where $r_N$ denotes the total number of relevant documents in the collection. And $I$-interpolated \emph{average precision} at recall level $x\in[0,1]$ for a single query $q$ is defined as:
\begin{equation*}
average\text{ }precision_q\equiv\frac{1}{I}\sum_{n=0}^{I-1}\hat{p}\left(\frac{n}{I-1}\right),
\end{equation*}
where as previously defined, $n$ denotes the $n$-th position in $\mathbf{r}$. We will use 11-point interpolated \emph{average precision} ($I=11$) as proposed in ref.~\cite{Kolda} since three out of four text collections used in our experiments are similar to those used in ref.~\cite{Kolda}. However, due to the dif\mbox{}ferences in the preprocessing steps, our results won't be similar to the results of ref.~\cite{Kolda}. And because there are several queries in each text collection (shown in the last row of table \ref{ch2:table16}), \emph{average precision} used in this work is the average value over \#Query. So, for each text collection:
\begin{equation*}
average\text{ }precision\equiv\frac{1}{Q}\sum_{q=1}^{Q}average\text{ }precision_q,
\end{equation*}
where $Q$ denotes \#Query.

\begin{figure}
 \begin{center}
  \subfigure[Medline]{
   \includegraphics[width=0.45\textwidth]{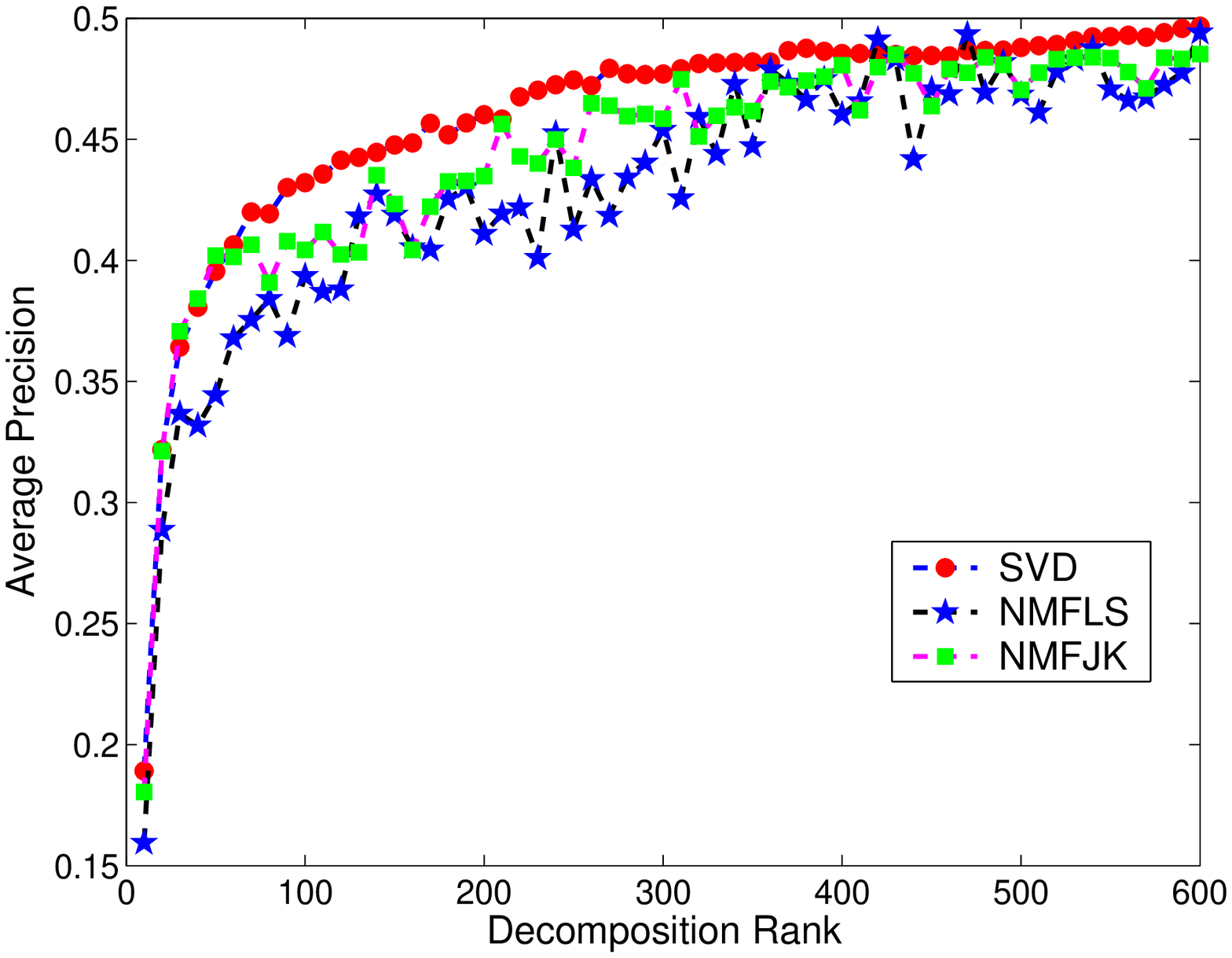}
   \label{ch3:fig3a}
  }
  \subfigure[Cranfield]{
   \includegraphics[width=0.45\textwidth]{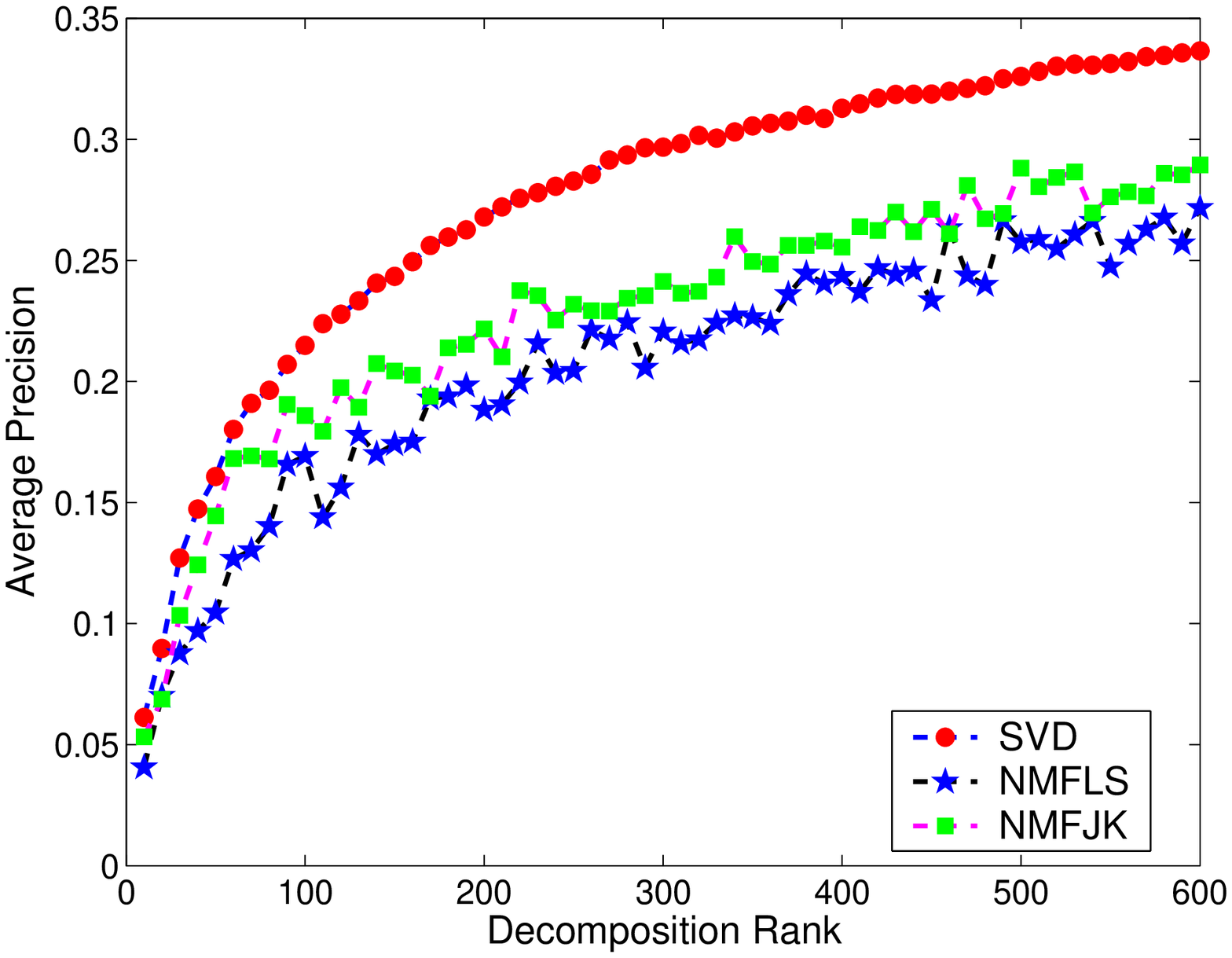}
   \label{ch3:fig3b}
  }
\\
  \subfigure[CISI]{
   \includegraphics[width=0.45\textwidth]{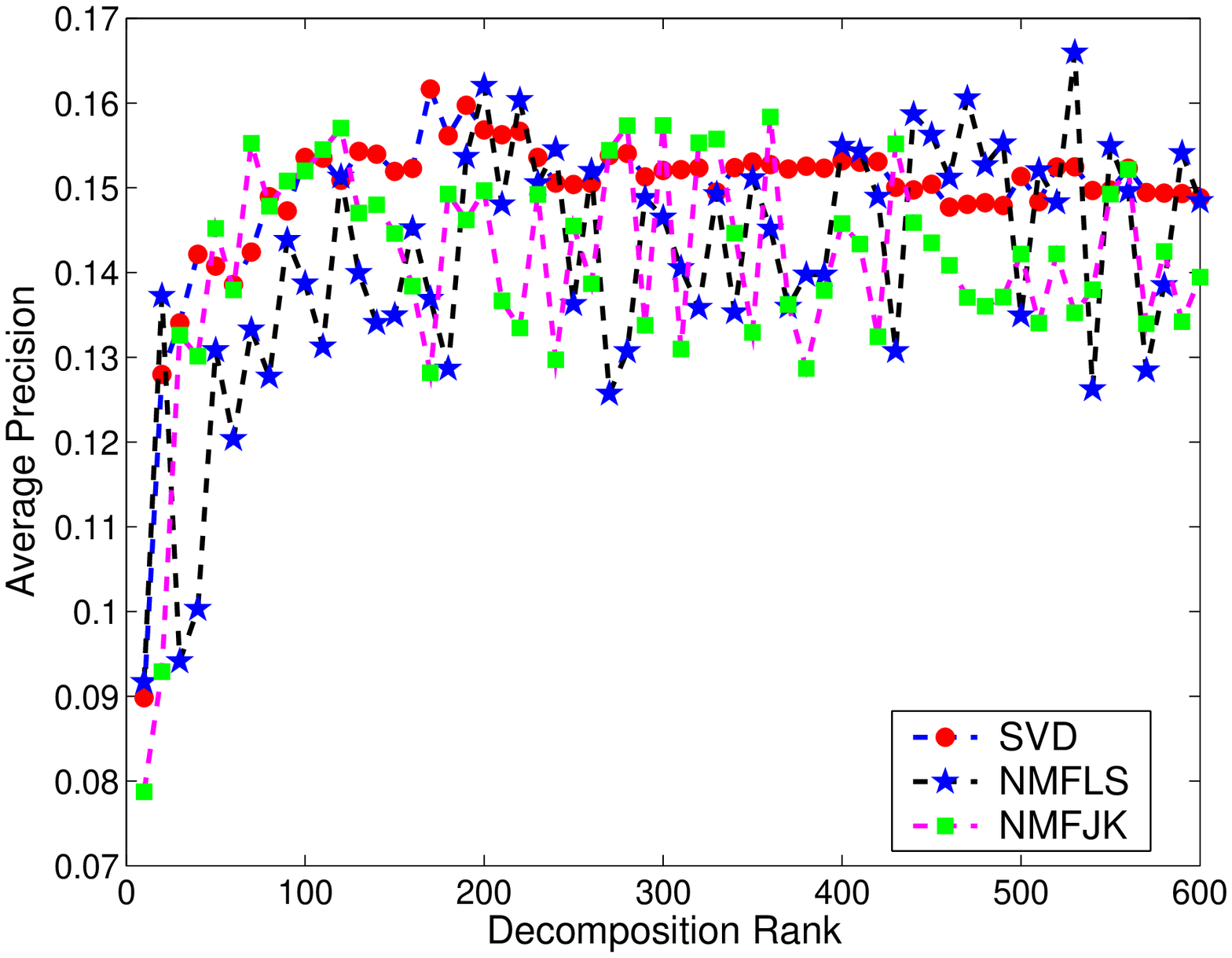}
   \label{ch3:fig3c}
  }
  \subfigure[ADI]{
   \includegraphics[width=0.45\textwidth]{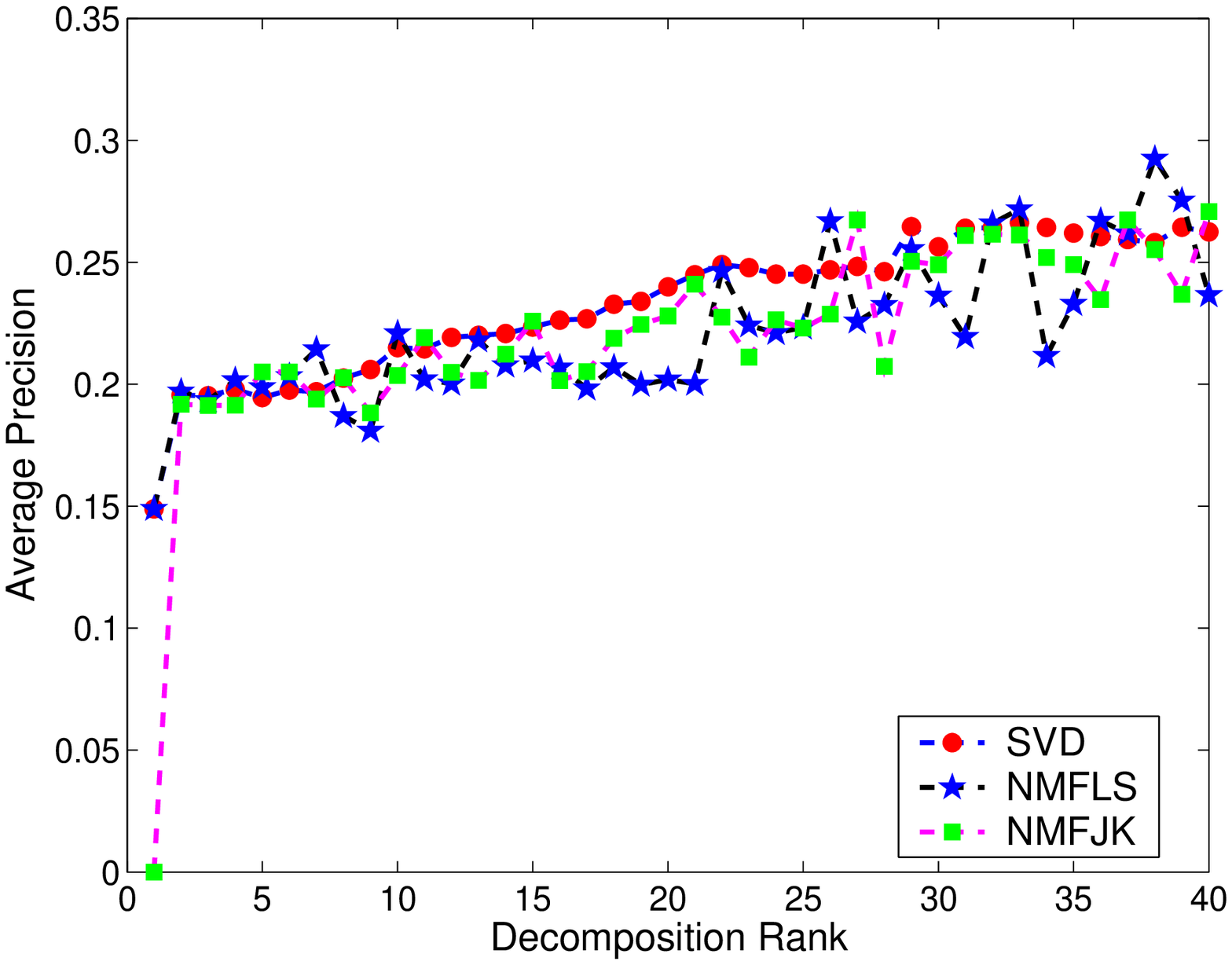}
   \label{ch3:fig3d}
  }
  \caption{{\bf Average precision} values over decomposition ranks.}
  \label{ch3:fig3}
 \end{center}
\end{figure}

\begin{table}[!ht]
 \begin{center}
   \caption{Average values of the {\bf average precision} over 10 trials.}
   \centering
   \begin{tabular}{|l|r|r|r|r|}
   \hline
             & Medline & Cranfield & CISI & ADI    \\
\hline
SVD & $\mathbf{0.4967}$(600) & $\mathbf{0.3365(600)}$ & $\mathbf{0.1617}$(170) & 0.2663(33) \\
NMFLS & 0.4769(600) & 0.2674(600) & 0.1510(530) & $\mathbf{0.2674(39)}$ \\
NMFJK & 0.4862(600) & 0.2871(600) & 0.1434(360) & 0.2610(40) \\
\hline 
  \end{tabular}
  \label{ch3:table12}
 \end{center}
\end{table}

Fig.~\ref{ch3:fig3} shows the \emph{average precision} values over decomposition ranks (for Medline, Cranfield, and CISI: [10, 20,$\ldots$,600], and for ADI: [1, 2,$\ldots$,40]) for all datasets. Table \ref{ch3:table12} displays average values of the \emph{average precision} over 10 trials with the values are in format \emph{val}(\emph{rank}), where \emph{val} denotes the average value over 10 trials at this \emph{rank}, and \emph{rank} denotes the rank where the maximum \emph{average precision} value is obtained at the first attempt (for example, in ADI at the first attempt the SVD reaches maximum \emph{average precision} at rank 33, NMFLS at rank 39, and NMFJK at rank 40, and this is also the ranks where peak values are achieved in fig.~\ref{ch3:fig3} for each dataset and each method). Note that because approximate matrices produced by the SVD are unique, there is no need to repeat the computation.

As shown in fig.~\ref{ch3:fig3}, in general the SVD produces better and more stable \emph{average precision} values over decomposition ranks for all datasets, with tendency the many the decomposition ranks the higher the \emph{average precision} values. The \emph{average precision} values produced by the NMF algorithms seem to be not stable, especially for CISI. NMFJK seems to have slightly better \emph{average precision} than NMFLS. This results are interesting since as discussed in section \ref{ch3:results1}, clustering capability of NMFJK is also better than NMFLS.

While there are cases in which NMFLS and NMFJK outperform the SVD, when the computations are repeated over 10 trials and the results are averaged, as shown in table \ref{ch3:table12}, the superiority of NMFLS and NMFJK seems to be vanished. This can be understood since NMF algorithms when converged, only stationarity of the limit points are guaranteed (so not even local-optimality is guaranteed by NMF algorithms). On the other hand, SVD algorithms not only have global-optimality guarantee, but also produce the same factors with dif\mbox{}ferences only in the numerical precision \cite{Kolda} (at least theoretically).

\begin{figure}
 \begin{center}
  \subfigure[Medline]{
   \includegraphics[width=0.45\textwidth]{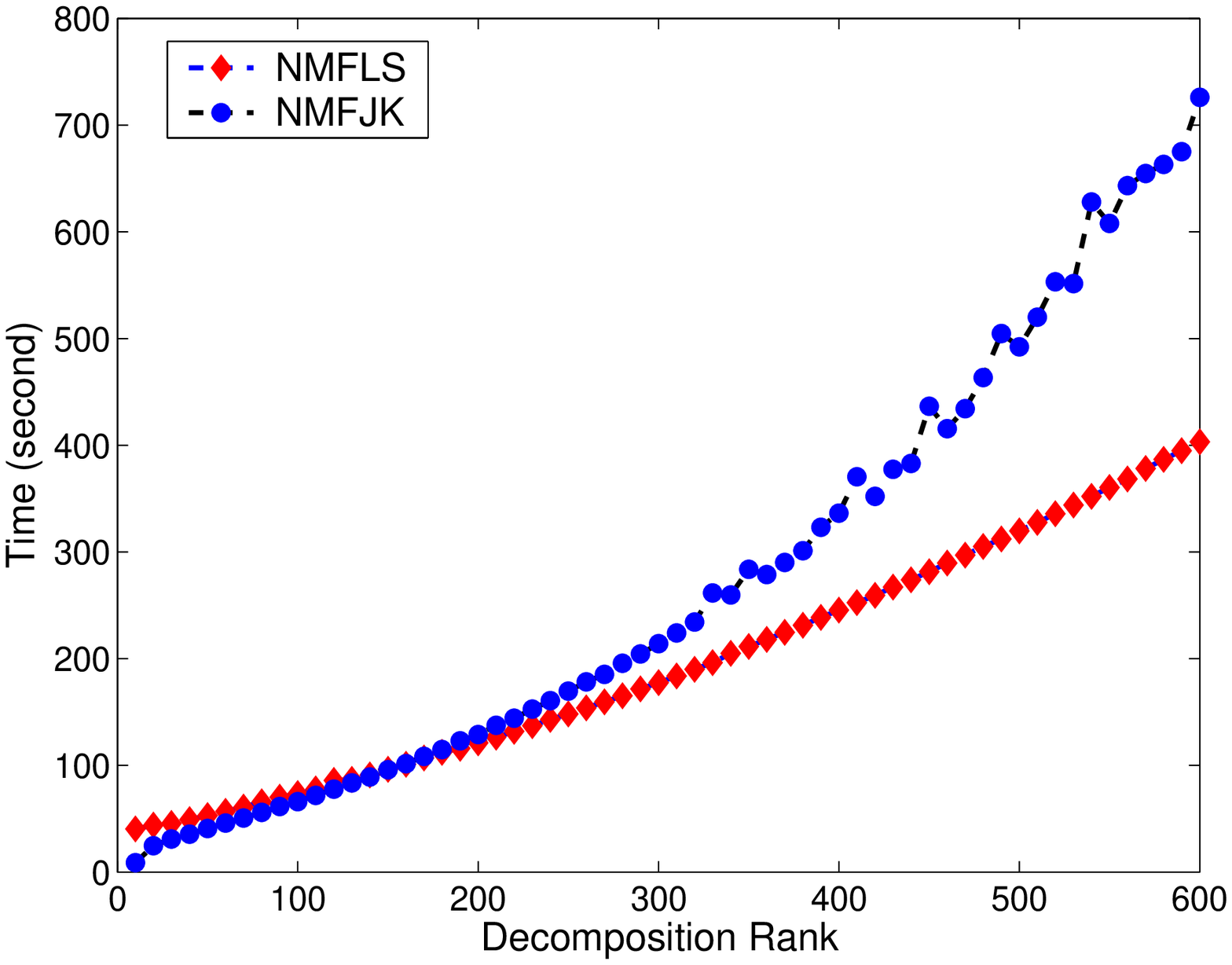}
   \label{ch3:fig4a}
  }
  \subfigure[Cranfield]{
   \includegraphics[width=0.45\textwidth]{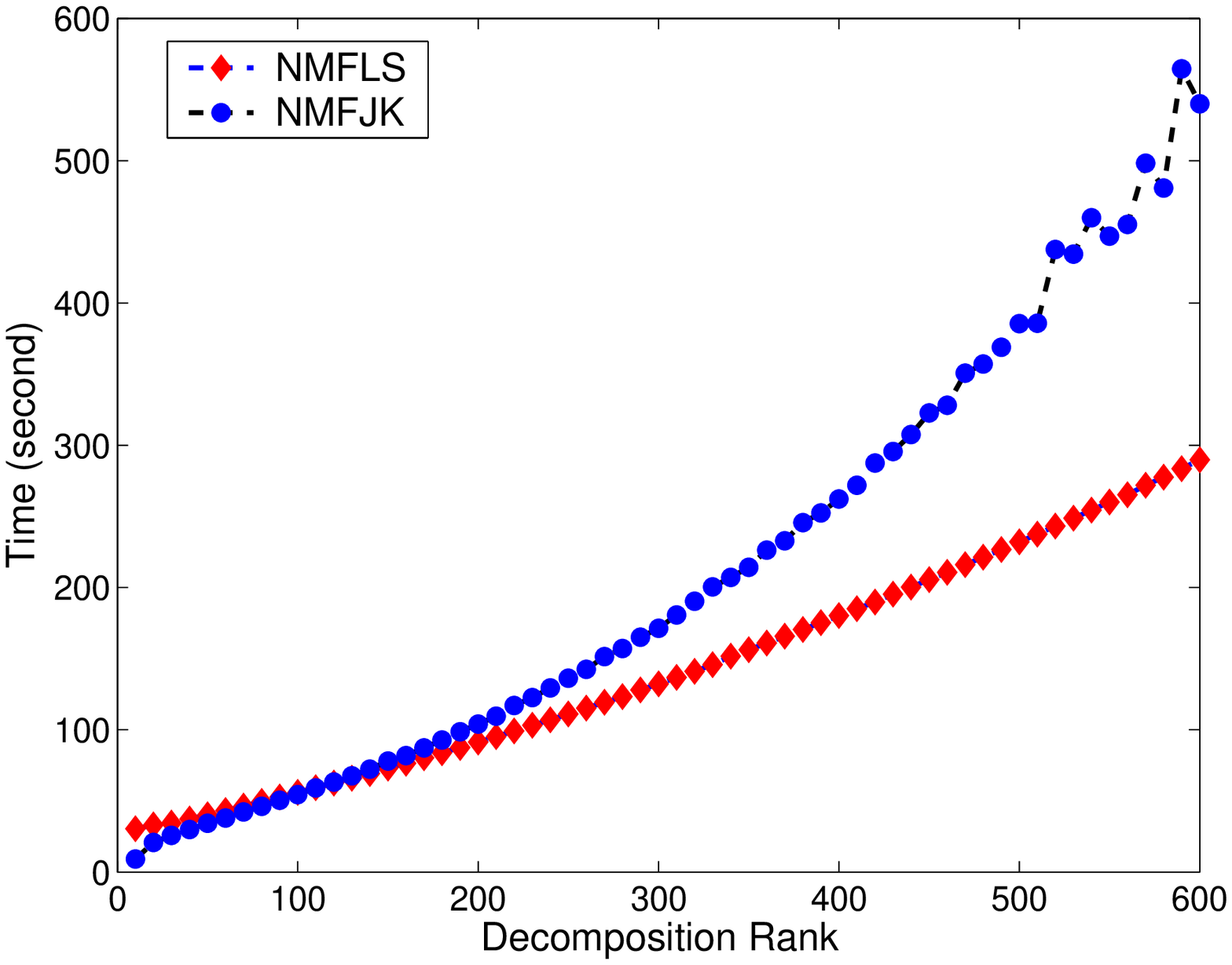}
   \label{ch3:fig4b}
  }
\\
  \subfigure[CISI]{
   \includegraphics[width=0.45\textwidth]{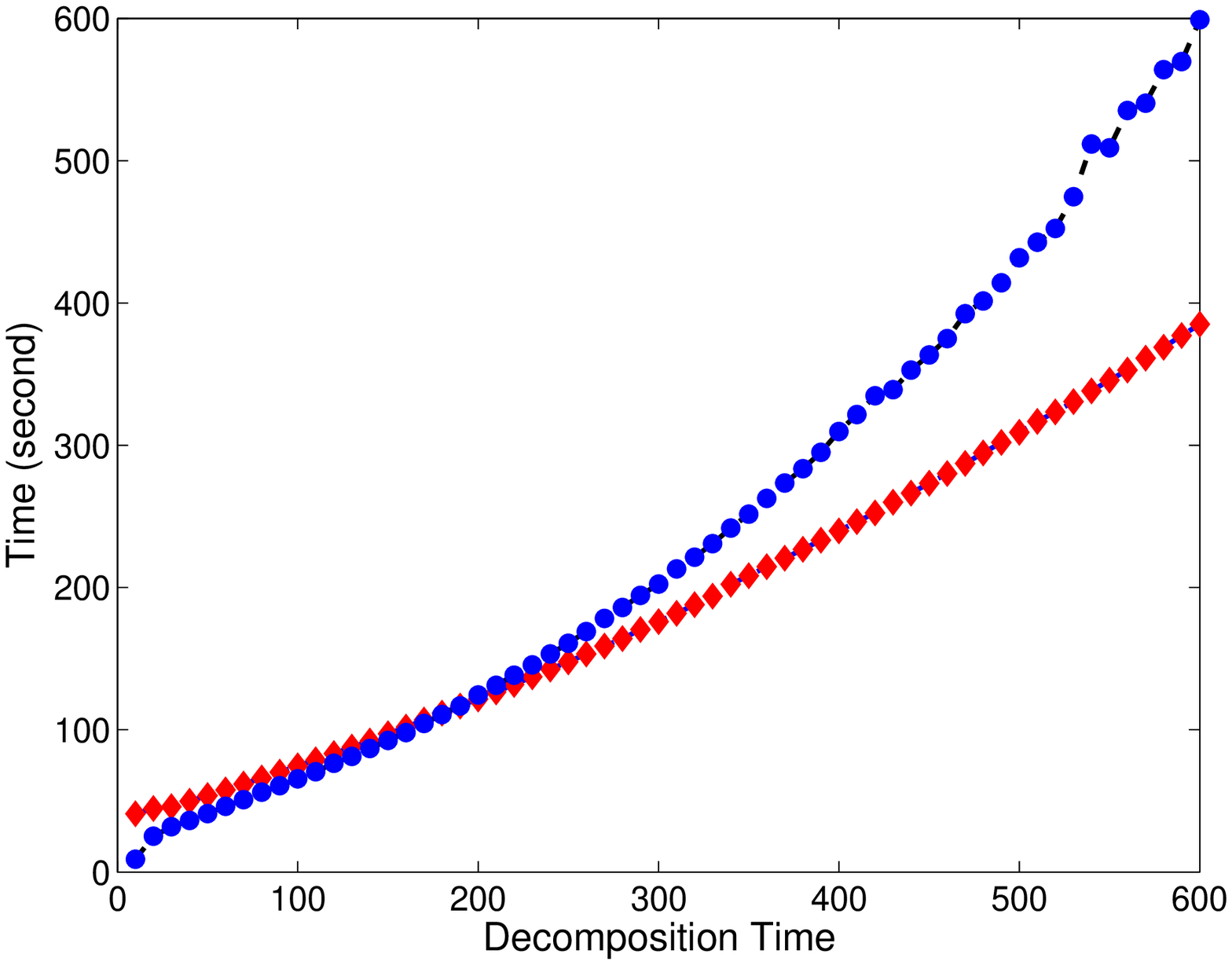}
   \label{ch3:fig4c}
  }
  \subfigure[ADI]{
   \includegraphics[width=0.45\textwidth]{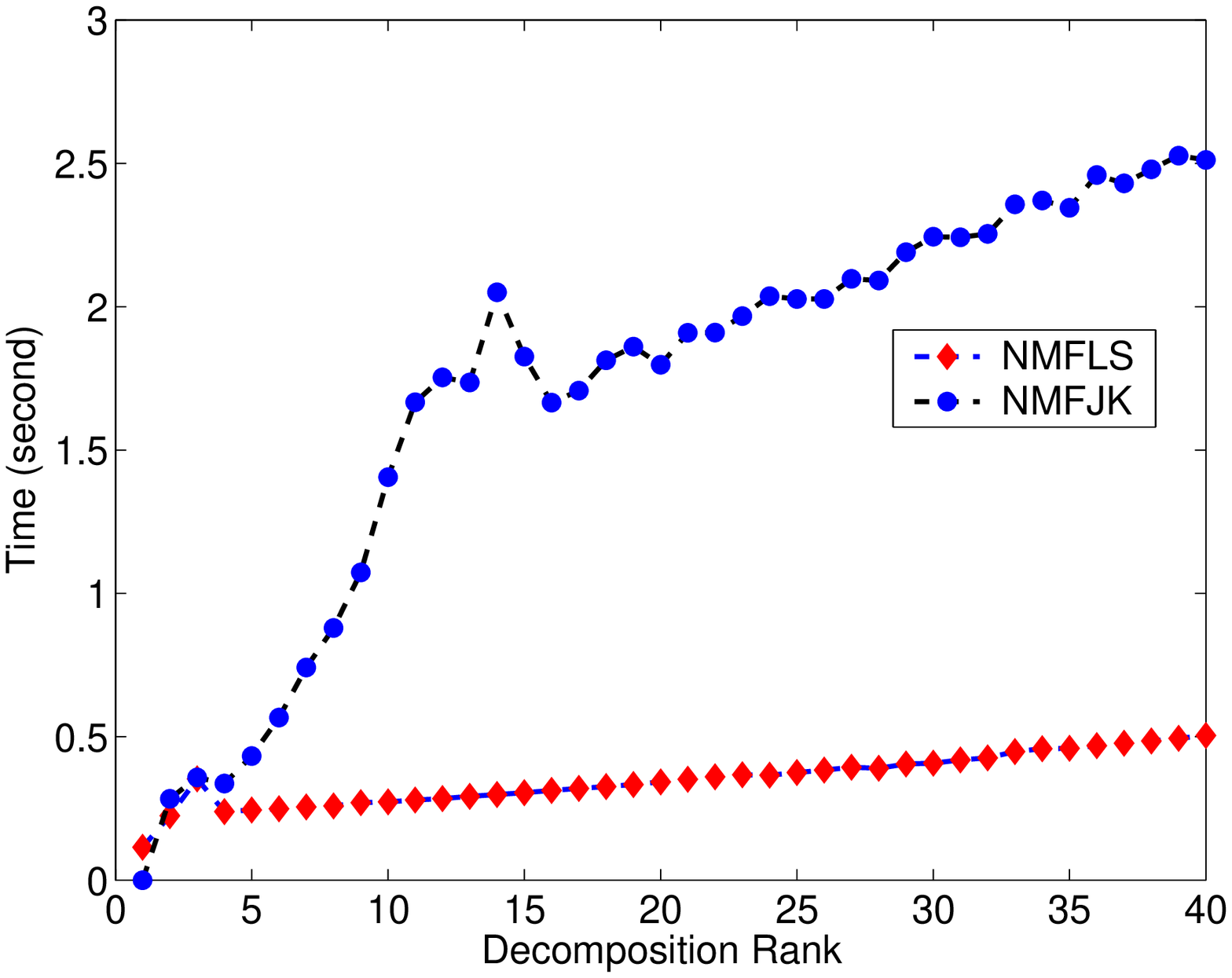}
   \label{ch3:fig4d}
  }
  \caption{Computational times over decomposition ranks (second).}
  \label{ch3:fig4}
 \end{center}
\end{figure}

Because rank-$k$ truncated SVD can be constructed from full rank SVD by taking the first $k$ columns of the singular matrices and the $k\times k$ principal submatrix of the singular value matrix, the computational times for the SVD are not recorded for each decomposition rank, rather we compute full rank SVD for each dataset and record the times which are $498.59$, $448.67$, $237.46$, and $1.1924$ seconds for Medline, Cranfield, CISI, and ADI respectively. And the computational times over decomposition ranks for NMFLS and NMFJK are shown in fig.~\ref{ch3:fig4}. 

As shown in fig.~\ref{ch3:fig4}, in general, for every dataset NMFJK is faster than NMFLS for lower ranks, but then the computational times of NMFJK are growing faster than NMFLS, resulting in slower performance for higher ranks. These results are interesting since according to the creator of NMFJK, this algorithm is the fastest NMF algorithm so far \cite{JKim2}. Table \ref{ch3:table5} can also be considered for evaluating the computational times of NMFJK which in the Reuters datasets, NMFJK is faster than NMFLS. However, since the decomposition ranks are rather very small (up to 12), the results in \ref{ch3:table5} are in accord to the results in fig.~\ref{ch3:fig4}. Thus, it seems that in lower ranks, NMFJK is faster than NMFLS, but in higher ranks NMFJK is slower than NMFLS. Table \ref{ch3:table13} shows the average computational times of NMFJK and NMFLS over 10 trials for decomposition ranks shown in table \ref{ch3:table12}. As the ranks are all high, NMFJK is slower than NMFLS for all datasets.

\begin{table}
 \begin{center}
   \caption{Average computational times over 10 trials.}
   \centering
   \begin{tabular}{|l|r|r|r|r|}
   \hline
             & Medline & Cranfield & CISI & ADI    \\
\hline
NMFLS & 367.8 & 283.5 & 158.6 & 0.4165 \\
NMFJK & 746.4 & 537.8 & 259.9 & 0.5045 \\
\hline 
  \end{tabular}
  \label{ch3:table13}
 \end{center}
\end{table}

\section{Conclusions} \label{ch3:conclusion}

We have presented a theoretical framework for supporting clustering aspect of the NMF without setting the KKT multipliers to zeros. Thus the stationary point used in proving this aspect is guaranteed to be on the nonnegative orthant which is the feasible region of the NMF. Our theoretical work implies a limitation of the NMF as a clustering method in which it cannot be used in clustering linearly inseparable datasets. So, the NMF as a clustering method is more resembling k-means clustering or SVM than the spectral clustering, even though both the NMF and the spectral methods utilize matrix decomposition techniques. As the clustering capabilities of k-means and SVM usually can be improved by using the kernel methods, probably the same approach can also be employed in the NMF. We will address this issue in our future researches.

Clustering capability of NMFJK is comparable to the SVD in Reuters datasets with NMFJK tends to be better for small \#cluster and the SVD for big \#cluster. But unfortunately, NMFLS which is the standard NMF algorithm cannot outperform the SVD. These results imply clustering aspect of the NMF is algorithm-dependent, a fact that seems to be overlooked in the NMF researches.

LSI aspect of the NMF seems to be comparable to the SVD in its power for solving synonymy and polysemy problems for datasets with clear semantic structures that allowed these problems to be revealed. In real datasets, however, the NMF generally cannot outperform the SVD. But an interesting fact comes into sight; in some cases, the NMF can outperform the SVD, even though when the computations are repeated and averaged over the number of trials, these advantages vanish. Because the NMF can offer different results depending on the algorithms, the initializations, the objectives, and the problems, improving LSI capability of the NMF is possible. We will address this problem in our future researches.

\end{document}